\documentclass{article}

\usepackage[preprint]{corl_2026} 
\usepackage{amsthm}
\usepackage{enumitem}
\usepackage[table]{xcolor}
\usepackage{threeparttable}
\usepackage{mathtools}
\usepackage{amsmath}
\usepackage{amsfonts}
\usepackage{wrapfig}
\usepackage{booktabs}       
\usepackage{algorithm}
\usepackage{algpseudocode}

\newtheorem{theorem}{Theorem}
\newtheorem{proposition}{Proposition}

\title{Efficient On-policy Visual-RL via Stochastic Decoupled Policy Gradient}

\author{
Haoxiang You\thanks{Equal contribution.}\; \thanks{Yale University, Department of Mechanical Engineering.}
\\
\texttt{haoxiang.you@yale.edu}
\And
Yilang Liu\footnotemark[1]\; \footnotemark[2]
\\
\texttt{yilang.liu@yale.edu}
\And
Davis Zong\thanks{Yale University, Department of Computer Science.}
\\
\texttt{davis.zong@yale.edu}
\And
Qian Wang\footnotemark[3]
\\
\texttt{peter.wang.qw262@yale.edu}
\And
Teeratham Vitchutripop\footnotemark[3]
\\
\texttt{tj.vitchutripop@yale.edu}
\And
Qi Wang\thanks{Shanghai Jiao Tong University, School of Computer Science.}
\\
\texttt{qiwang067@sjtu.edu.cn}
\And
Daniel Rakita\footnotemark[3]
\\
\texttt{daniel.rakita@yale.edu}
\And
Ian Abraham\footnotemark[2]\;\thanks{University of Sydney, School of Electrical and Computer Engineering.}
\\
\texttt{ian.abraham@sydney.edu.au}
}

\begin{document}
\maketitle


\begin{abstract}

We present the stochastic decoupled policy gradient (SDPG), a lightweight visual reinforcement learning (RL) method that trains diverse visuomotor control policies end-to-end within a few hours on a single NVIDIA RTX 4080 GPU.
SDPG estimates policy gradients via random perturbations of trajectory rollouts, requiring orders of magnitude fewer batch-rendered environments and substantially reducing compute and memory overhead.
On visual MuJoCo benchmarks, SDPG consistently outperforms baseline methods in training time, memory usage, and rewards.
Finally, to support future research, we introduce a suite of realistic visual robotics benchmarks spanning dexterous manipulation, challenging locomotion, and demonstrate effective sim-to-real transfer on physical hardware.
Videos are available at
\href{https://haoxiangyou.github.io/sdpg-website/}{https://haoxiangyou.github.io/sdpg-website/}.
\end{abstract}



\section{Introduction} \label{sec:intro}
\begin{wrapfigure}{h}{0.48\textwidth}
    \vspace{-50pt}
    \begin{center}
    \includegraphics[width=0.48\textwidth]{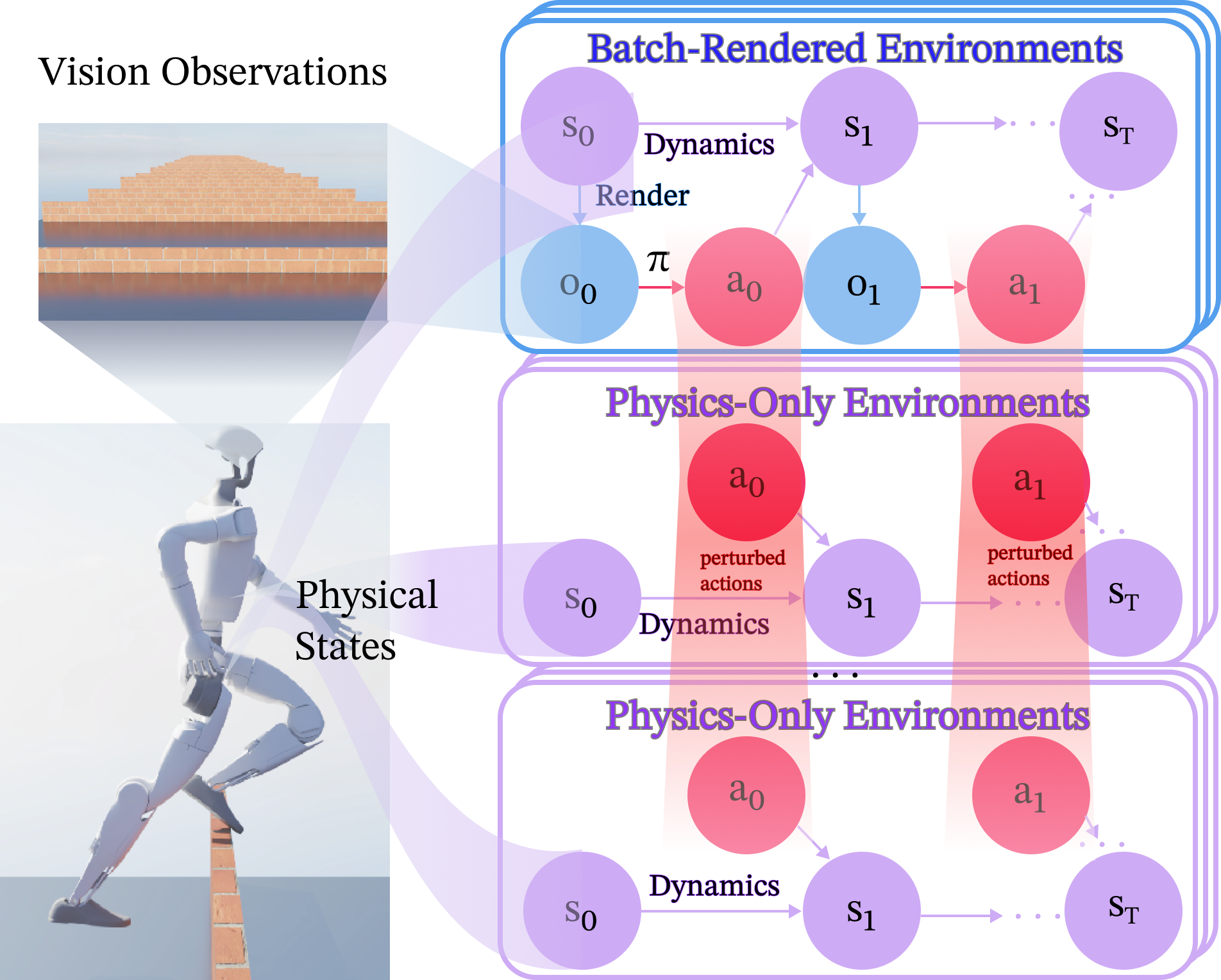}
    \end{center}
    \vspace{-10pt}
    \caption{\small{SDPG combines batch-rendered and physics-only environments to estimate policy gradients. Batch-rendered environments evaluate policy performance, while the physics-only environments provide perturbed rollouts for policy improvement.}}
    \vspace{-15pt}
    \label{fig:memory_scaling}
\end{wrapfigure}
Learning control policies directly from visual inputs is a central challenge in robotics, enabling applications ranging from autonomous navigation to dexterous manipulation. 
However, visual reinforcement learning (RL) is significantly more resource-demanding than learning from low-dimensional states. 
Existing approaches such as DrQV2~\citep{yarats2021masteringvisualcontinuouscontrol}, TD-MPC2~\citep{hansen2024tdmpc2}, and DreamerV3~\citep{hafner2024masteringdiversedomainsworld} require extensive sequential updates to learn effective representations, resulting in slow wall-clock training.
On the other hand, on-policy methods such as PPO rely on large numbers of parallel environments with batch rendering~\citep{tao2025maniskill3gpuparallelizedrobotics, zakka2025mujocoplayground}, which is highly memory-intensive and difficult to scale without GPU clusters~\citep{nvidia2025isaaclabgpuacceleratedsimulation, singh2025endtoendrlimprovesdexterous, tao2025maniskill3gpuparallelizedrobotics}.
An alternative approach is simulation-based distillation, where a low-dimensional teacher policy is first trained in simulation and a visual student policy then learns by imitating the teacher with DAgger~\citep{Lee_2020,Miki_2022,agarwal2022leggedlocomotionchallengingterrains,lum2024dextrahgpixelstoactiondexterousarmhand}.
While efficient, this approach often yields suboptimal performance~\citep{rudin2025parkourwildlearninggeneral, singh2025endtoendrlimprovesdexterous} due to: \emph{information asymmetry}, where the student lacks access to privileged state information~\citep{2014-mfcgps, kim2025distillingrealizablestudentsunrealizable}, and \emph{distributional shift}, where compounding imitation errors lead to unrecoverable failures~\citep{singh2025endtoendrlimprovesdexterous, you2025accelerating}.

Recently, several works combine differentiable simulation and first-order RL to train visual policies~\citep{you2025accelerating, pan2026learningflyrapidpolicy, Zhang_2025}. 
These approaches achieve lower-variance gradient estimates with fewer samples and enable continuous optimization for asymptotic performance.
Building on this, \citet{you2025accelerating} further introduce \emph{decoupled policy gradients}, which improve training stability and efficiency via a stop gradient trick, achieving superior performance in training time, memory usage and rewards on Mujoco benchmarks.
However, the practical adoption of first-order RL remains limited due to several challenges. First, backpropagating through long-horizon dynamics causes severe gradient instability, especially in contact-rich manipulation tasks. 
Second, these methods rely on full trajectory differentiation, which reduces physical fidelity due to soft contact models~\citep{schwarke2025learningdeployablelocomotioncontrol, song2024learningquadrupedlocomotionusing} and restricts the use of non-differentiable reward components~\citep{Hwangbo_2019, rudin2022learningwalkminutesusing}, which are often crucial for closing the sim-to-real gap.
Finally, many popular robotics simulators~\citep{isaacsim, tao2025maniskill3gpuparallelizedrobotics,mujoco_warp,Genesis} provide limited support for differentiable dynamics and sensor models, further limiting flexibility.

To address these practical challenges, we propose \emph{Stochastic Decoupled Policy Gradient (SDPG)}, an efficient RL method for training visual policies end-to-end. 
Instead of computing trajectory gradients analytically, SDPG estimates gradients via randomly perturbations, eliminating the need for full trajectory differentiation. 
We further introduce an adaptive exploration strategy and reward-invariant  normalization, which together yield numerically stable updates throughout training.
Beyond the algorithm, we introduce a unified perspective on policy gradient methods, connecting classical approaches and providing practical guidance for tuning visual RL.
We evaluate SDPG on visual MuJoCo benchmarks and find that it achieves highest reward while being memory and wall-clock time efficient. 
To facilitate further research, we provide a suite of egocentric benchmark tasks, spanning dexterous manipulation and challenging locomotion. 
Finally, we demonstrate sim-to-real transfer by deploying the learned policy on a Unitree Go2 robot.

\textbf{In summary, our contributions are}:
(a) We propose SDPG, a wall-clock and memory-efficient method for visual RL;
(b) We provide a unified perspective on policy-based methods;
(c) We introduce a suite of realistic visual RL benchmarks and demonstrate sim-to-real transfer capability.

\begin{figure}[t]
    \vspace{-15pt}
    \begin{center}
    \includegraphics[width=\textwidth]{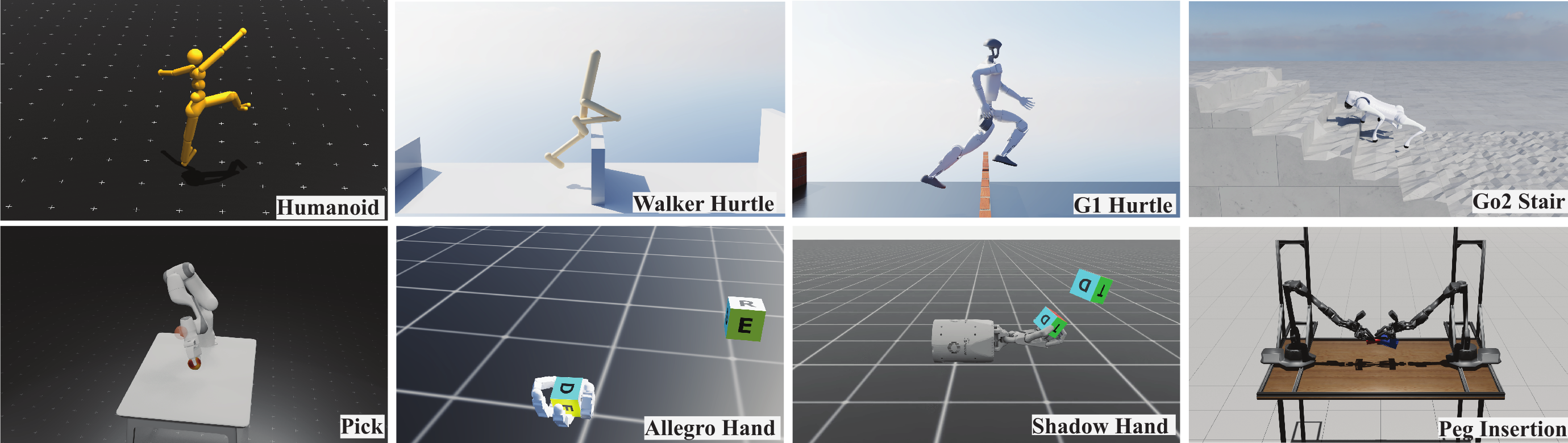}
    \end{center}
    \vspace{-10pt}
    \caption{\textbf{Preview of tasks:} including both manipulation and locomotion, egocentric and third-person view, single and multiple cameras, RGB and depth image. Our method learns all tasks end-to-end on a single NVIDIA RTX 4080 GPU within a few hours.}
    \vspace{-20pt}
    \label{fig:preview}
\end{figure}
\section{Preliminary} \label{sec: preliminary}
\subsection{Problem formulation}
In this work, we consider a dynamical system $\mathbf{s}_{t+1} = f(\mathbf{s}_t, \mathbf{a}_t)$, where $\mathbf{s}_t \in \mathcal{S}$ is the state and $\mathbf{a}_t \in \mathcal{A}$ is the control.
Let $\mathbf{o}_t = g(\mathbf{s_t})$ be the observation, where $g: \mathcal{S} \rightarrow \mathcal{O}$ denotes the sensor model, e.g.\ cameras, IMUs.
A policy $\mathbf{\pi}(\cdot|\mathbf{o}_t, \boldsymbol{\theta}): \mathcal{O} \times \Theta \rightarrow \Delta(\mathcal{A})$ maps observations to action distributions.
Given an initial condition $\mathbf{s}_0$, one can generate a trajectory $\tau = \{\mathbf{s}_0, \mathbf{a}_0, \dots \mathbf{s}_T, \mathbf{a}_T\}$ by rolling out the policy. 
Let the return be $\mathcal{J} (\tau) = \sum_{t=0}^T \gamma^t R(\mathbf{s}_t, \mathbf{a}_t)$, where $R : \mathcal{S} \times \mathcal{A} \rightarrow \mathbb{R}$ is the reward function and $\gamma \in [0, 1)$ is the discount factor.
The goal of policy optimization is to maximize the expected return under given initial distribution:
$
\boldsymbol{\theta}^\star
= \arg\max_{\boldsymbol{\theta}}\;
\mathcal{V}(\boldsymbol{\theta})
\coloneqq
\mathbb{E}_{\mathbf{s}_0 \sim \rho_0}
\big[ \mathcal{J}(\mathbf{s}_0, \boldsymbol{\theta}) \big]$.

\subsection{First-order RL and decoupled policy gradient}

\paragraph{First-order RL}

First-order RL estimates the policy gradient by differentiating through trajectories:
$\nabla_{\boldsymbol{\theta}} \mathcal{V} = \mathbb{E}_{\tau} [\nabla_{\boldsymbol{\theta}} \mathcal{J} (\tau)] \coloneqq \mathbb{E}_{\tau } \Big[\frac{d\mathbf{A}}{d\boldsymbol{\theta}}^\top \nabla_\mathbf{A} \mathcal{J} \Big]
$,
where $\mathbf{A} \coloneqq \{\mathbf{a}_0, \dots, \mathbf{a}_T\}$ denotes the sequence of actions, $\nabla_{\mathbf{A}} \mathcal{J}$ is the gradient of the return with respect to the action sequence, and $\frac{d\mathbf{A}}{d\boldsymbol{\theta}}$ denotes the total derivative of the action sequence with respect to the policy parameters.

\paragraph{Decoupled policy gradient}
\citet{you2025accelerating} introduce a decoupling trick that stops gradients through the observations: $\mathbf{a}_t = \pi(\cdot \mid \text{sg}(\mathbf{o}_t), \boldsymbol{\theta})$
, resulting in the \emph{decoupled policy gradient} $\nabla_{\boldsymbol{\theta}} \mathcal{V}_{\text{decoupled}} \coloneqq \mathbb{E}_{\tau } \Big[\frac{\partial\mathbf{A}}{\partial\boldsymbol{\theta}}^\top \nabla_\mathbf{A} \mathcal{J} \Big]$, where total derivative $\frac{d\mathbf{A}}{d\boldsymbol{\theta}}$ is replaced by $\frac{\partial\mathbf{A}}{\partial\boldsymbol{\theta}}$.
This formulation avoids backpropagating through the rendering pathway, significantly reducing the computational and memory cost associated with images.
However, two key challenges still remain. 
First, gradient estimation can become unstable or explode due to long computation chains~\citep{metz2022gradientsneed,suh2022differentiablesimulatorsbetterpolicy}, especially in contact-rich settings. 
Second, these methods require differentiability through entire trajectories, which is not well-supported in common robotics simulators and leads to brittle sim-to-real design.

\section{Method} \label{sec:method}

We propose the stochastic decoupled policy gradient (SDPG) to address gradient instability and eliminate the requirement of full trajectory differentiation. 
We first present the core idea, then provide a unified view connecting our method to existing policy gradient approaches.

\subsection{Stochastic Decoupled Policy Gradient}

Our key idea is to replace the analytical trajectory Jacobian $\nabla_{\mathbf{A}} \mathcal{J}$ with a smoothing estimation.
Concretely, for each action sequence $\mathbf{A}$ from current policy $\pi_{\boldsymbol{\theta}}$, we generate $M$ random perturbations ${\mathcal{E}^j \coloneqq [\boldsymbol{\epsilon}_0^j, \dots, \boldsymbol{\epsilon}_T^j]}, j=1,\dots, M$, where $\boldsymbol{\epsilon}_t^j \sim \mathcal{N}(\mathbf{0}, \mathbf{I})$, and evaluate the corresponding returns.
The trajectory gradient is then estimated as the weighted average:
\begin{equation}
     \nabla \mathcal{J}_\text{smooth} (\mathbf{A}) \coloneqq \mathbb{E}_{\mathcal{E} \sim \mathcal{N}(\mathbf{0},\mathbf{I})}\Big[\frac{ \mathcal{J}(\mathbf{A} + \delta \mathcal{E}) - \mathcal{J}(\mathbf{A})}{\sigma} \mathcal{E}\Big] \approx \frac{1}{M}\sum_{j=1}^M \Big( \frac{ \mathcal{J}(\mathbf{A} + \delta \mathcal{E}^j) - \mathcal{J}(\mathbf{A})}{\sigma} \mathcal{E}^j \Big), \label{eq:smoother_action_improved_direction}
\end{equation}
where $\delta$ is the exploration factor and $\sigma$ is the standard deviation of returns.

The estimator $\nabla \mathcal{J}_{\text{smooth}}(\mathbf{A})$ can be interpreted as the gradient of a smoothed surrogate of the objective $\mathcal{J}(\mathbf{A})$. 
In particular, the following result holds:

\begin{theorem}\label{thm:stochastic_smoothing}(See Appendix~\ref{sec: proof of thm stochastic} for proof)
Let $\mathcal{J}_\delta (\mathbf{A}) \coloneqq\mathbb{E}_{\mathcal{E} \sim \mathcal{N}(\mathbf{0}, \mathbf{I})}[\mathcal{J}(\mathbf{A} + \delta \mathcal{E})]$ denote the return expectation under Gaussian noise. 
Then, $\nabla \mathcal{J}_\text{smooth} (\mathbf{A}) = \frac{\delta}{\sigma} \nabla_\mathbf{A} \mathcal{J}_\delta (\mathbf{A}) $. 
\end{theorem}

\begin{wrapfigure}{h}{0.485\textwidth}
    \vspace{-20pt}
    \begin{center}
    \includegraphics[width=0.485\textwidth]{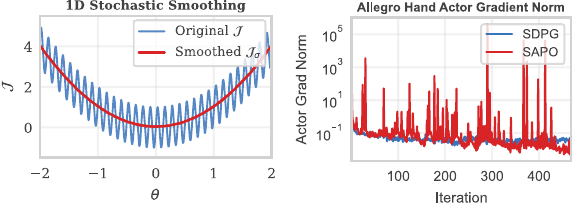}
    \end{center}
    \vspace{-15pt}
    \caption{\small{\textbf{Left:} A 1D toy example showing the original function and its smoothed surrogate. \textbf{Right:} Gradient norms during training for SAPO~\citep{xing2025stabilizing} and SDPG (Ours). SDPG maintains stable gradients, while SAPO exhibits large spikes.}}
    \vspace{-15pt}
    \label{fig:stochastic_smoothing}
\end{wrapfigure}

Replacing the true gradient $\nabla_{\mathbf{A}} \mathcal{J}$ with the smoothed estimator $\nabla \mathcal{J}_{\text{smooth}} (\mathbf{A})$ offers several benefits.
First, it yields smoother updates. 
As shown in Figure~\ref{fig:stochastic_smoothing}, $\nabla \mathcal{J}_{\text{smooth}} (\mathbf{A})$ provides smoother gradients, whereas the analytical gradient exhibits large spikes.
Second, it removes the need for differentiability, enabling seamless use with existing simulators.
Third, it avoids backward passes, reducing computation time by half.

We now discuss how to turn this update direction into a policy improvement step. 
To this end, we introduce a technique that will be used throughout this work.
\begin{proposition}\label{prop: update rule and supervised trick}(See Appendix~\ref{sec: proof of prop} for proof)
Given an update rule of the form $\boldsymbol{\theta}_{k+1} \leftarrow \boldsymbol{\theta}_k + \beta {\frac{\partial \mathbf{A}_k}{\partial\boldsymbol{\theta}_k}}^\top \mathbf{d}(\mathbf{A}_k)$,
where  $\mathbf{A}_k = \pi(\boldsymbol{\theta}_k)$ is an intermediate variable, and $\mathbf{d}(\mathbf{A}_k)$ is an update direction, e.g., $\nabla_{\mathbf{A}_k} \mathcal{J}$,
we can equivalently express the update as a gradient descent step on the following supervised objective:
\begin{equation}
\boldsymbol{\theta}_{k+1} \leftarrow \boldsymbol{\theta}_k - \beta \nabla_{\boldsymbol{\theta}_k} \mathcal{L}, \quad \text{where} \quad
\mathcal{L} \coloneqq \tfrac{1}{2}\|\mathbf{A}_k - \mathbf{A}_k^{\mathrm{target}}\|^2, \ \mathbf{A}_k^{\mathrm{target}} \coloneqq \mathrm{sg}\big(\mathbf{A}_k + \mathbf{d}(\mathbf{A}_k)\big)
\end{equation}
where $\mathrm{sg}(\cdot)$ denotes the stop-gradient operator.
\end{proposition}

Using Proposition~\ref{prop: update rule and supervised trick}, we update the policy by optimizing the following objective:
\begin{equation}
\mathcal{L}(\boldsymbol{\theta}, \mathbf{O}, \mathbf{A}) \coloneqq \mathbb{E} \left[ \| \pi(\mathbf{O} \mid \boldsymbol{\theta}) - \mathrm{sg}\big(\mathbf{A} + \nabla \mathcal{J}_\text{smooth} (\mathbf{A})\big) \||_2^2 \right] \label{eq: stochastic bc loss},
\end{equation}
where $\mathbf{O} \coloneqq\{\mathbf{o}_0,\dots \mathbf{o}_T \}$ are the observations from rollouts.
The \emph{stochastic decoupled policy gradient} (SDPG) is defined as 
$\nabla_{\boldsymbol{\theta}} \mathcal{V}_\text{SDPG} \coloneqq -\nabla_{\boldsymbol{\theta}} \mathcal{L}(\boldsymbol{\theta}, \mathbf{O}, \mathbf{A})$.
Substituting Equation~\eqref{eq:smoother_action_improved_direction} into~\eqref{eq: stochastic bc loss}, we can write the SDPG as:
\begin{equation}
\begin{aligned}
\nabla_{\boldsymbol{\theta}} \mathcal{V}_\text{SDPG} 
&\coloneqq \mathbb{E} \Big[\frac{\partial \mathbf{A}}{\partial \boldsymbol{\theta}}^\top \nabla \mathcal{J}_\text{smooth}(\mathbf{A})\Big] = \mathbb{E} \Big[
\frac{\partial \mathbf{A}}{\partial \boldsymbol{\theta}}^\top 
\mathbb{E}_{\mathcal{E} \sim \mathcal{N}(\mathbf{0},\mathbf{I})}
\big[
\frac{\mathcal{J}(\mathbf{A} + \delta \mathcal{E}) - \mathcal{J}(\mathbf{A})}{\sigma} \, \mathcal{E}
\big]
\Big] \\
&\approx \mathbb{E}_{\tau \sim \pi(\cdot|\boldsymbol{\theta}), \mathbf{s}_0 \sim \rho_0} \Big[
\frac{1}{M} \sum_{j=1}^M \frac{\partial \mathbf{A}}{\partial \boldsymbol{\theta}}^\top 
\frac{\mathcal{J}(\mathbf{A} + \delta \mathcal{E}^j) - \mathcal{J}(\mathbf{A})}{\fcolorbox{red}{white}{$\sigma$}} \, \mathcal{E}^j
\Big]
\end{aligned}.
\label{eq: sdpg derivation}
\end{equation}

\subsection{A Unified View of Policy Gradient Methods}

Here, we show how SDPG relates to classical policy gradient methods and present an alternative perspective that unifies policy-based methods.

\paragraph{Connection to policy gradient method}

Recall the policy gradient estimator:
$\nabla_{\boldsymbol{\theta}} \mathcal{V}_\text{PG} = \mathbb{E} \big[(\mathcal{J} - \mathcal{B}) \sum_{t=0}^T \nabla_{\boldsymbol{\theta}} \log \pi_{\boldsymbol{\theta}} (\mathbf{u}_t|\mathbf{o}_t) \big]$,
where $\mathbf{u}_t$ is an action sampled from $\pi_{\boldsymbol{\theta}}$, $\mathcal{J}$ denotes the return of the trajectory and $\mathcal{B}$ is a baseline.
For continuous control, the policy is typically parameterized as a Gaussian.
In this case, we write the action as $\mathbf{u}_t = \mathbf{a}_t + \delta \epsilon_t$, where $\mathbf{a}_t = \mu_{\boldsymbol{\theta}} (\mathbf{o}_t)$ is the mean, $\delta$ is the standard deviation and $\epsilon_t$ is noise.
The return $\mathcal{J}$ is a function of the action sequence ${\mathbf{u}_t}$ and can therefore be written as $\mathcal{J}(\mathbf{U}) \coloneqq\mathcal{J}(\mathbf{A} + \delta \mathcal{E})$. 
If we choose the baseline $\mathcal{B} = \mathcal{J}(\mathbf{A})$ as the return of mean action and substitute Gaussian policy into $\nabla_{\boldsymbol{\theta}} \log \pi_{\boldsymbol{\theta}} (\mathbf{u}_t \mid \mathbf{s}_t)$, we obtain the policy gradient as following:
\begin{equation}
    \nabla_{\boldsymbol{\theta}} \mathcal{V}_\text{PG} = \mathbb{E}_{\tau \sim \pi(\cdot|\boldsymbol{\theta}), \mathbf{s}_0 \sim \rho_0}\Big[ \frac{\partial \mathbf{A}}{\partial \boldsymbol{\theta}} ^\top \frac{\mathcal{J}(\mathbf{A} + \delta \mathcal{E}) - \mathcal{J}(\mathbf{A})}{\fcolorbox{red}{white}{$\delta$}} \mathcal{E}\Big] \label{eq: reinforce under gaussian}.
\end{equation}

\begin{wrapfigure}{h}{0.5\textwidth}
    \vspace{-20pt}
    \begin{center}
    \includegraphics[width=0.50\textwidth]{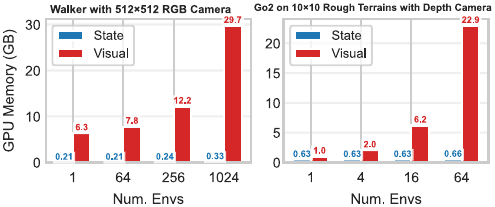}
    \end{center}
    \vspace{-15pt}
    \caption{\small{\textbf{Memory scaling.} Visual-based environments require significantly more memory than state-based environments. \textbf{Left:} Walker with an RGB sensor using GS-Madrona as the rendering backend. \textbf{Right:} Go2 traversing diverse terrains with a depth camera, using a BVH-based rendering backend.}}
    \vspace{-10pt}
    \label{fig:memory_scaling}
\end{wrapfigure}
In practice, the expectation is estimated via Monte Carlo by averaging over $N$ trajectories, which we refer to as \emph{nominal trajectories}. 
Comparing Equation~\eqref{eq: sdpg derivation} and Equation~\eqref{eq: reinforce under gaussian}, we see that the policy gradient~\eqref{eq: reinforce under gaussian} is a special case of SDPG~\eqref{eq: sdpg derivation} where only one perturbation is sampled to estimate the gradient, i.e.\ $M=1$ \footnote{And the constant $\delta$ in the denominator is replaced by $\sigma$ (marked by red boxes) for numerical stability.}.
Although the difference appears minor, it yields practical gains in memory-intensive visual RL.
Vanilla policy gradient relies on a large number of nominal trajectories $N$, requiring batched rendering across all trajectories.
In contrast, SDPG estimates gradients using multiple perturbations on the same nominal trajectories, avoiding repeated policy evaluations, visual processing, and achieving reliable gradient estimates with substantially fewer batched renderings, thus saving memory usage.

\paragraph{Policy gradient from a guided search perspective}
Here, we present an alternative perspective on policy-based methods. 
While it does not change the algorithm, we find this perspective useful for guiding hyperparameter tuning and other design choices.

We begin by showing how the decoupled policy gradient connects to another class of methods: guided search~\citep{pmlr-v28-levine13,gu2016continuousdeepqlearningmodelbased,chebotar2018pathintegralguidedpolicy}, which first performs local trajectory optimization and then distills the results into a policy.
Consider an action sequence $\mathbf{A} \coloneqq [\mathbf{a}_0, \dots \mathbf{a}_T]$ from the current policy.
We can improve the action sequence via gradient ascent:
\begin{equation}
    \mathbf{A}^\text{target} = \mathbf{A}+ \alpha \nabla_{\mathbf{A}} \mathcal{J} (\mathbf{A}); \  \nabla_{\mathbf{A}} \mathcal{J} \coloneqq \{\nabla_{\mathbf{a}_0} \mathcal{J}, \dots, \nabla_{\mathbf{a}_T} \mathcal{J} \}. \label{eq: traj_opt_grad_ascent}
\end{equation}
To utilize these trajectories for policy improvement, we construct a supervised behavior cloning objective:
\begin{equation}
    \mathcal{L}_\text{BC}(\boldsymbol{\theta}, \mathbf{O}, \mathbf{A}^\text{target}) \coloneqq \mathbb{E} \Big[\frac{1}{2\alpha } \sum_{t=0}^T \|\pi(\mathbf{o}_t, \boldsymbol{\theta})  - \mathbf{a}_t^\text{target} \|_2^2 \Big]. \label{eq:behavior_cloning_loss}
\end{equation}
Then, we have the following result~(originally shown by~\citet{you2025accelerating}):
\begin{theorem} \label{theorem:connection_between_pg_and_gps}(Proof in Appendix~\ref{sec: proof of pg and gps})
The decoupled policy gradient equals the negative gradient of the behavior cloning loss~\eqref{eq:behavior_cloning_loss}, i.e., $\nabla_{\boldsymbol{\theta}} \mathcal{V}_{\text{decoupled}} \coloneqq \mathbb{E}_{\tau } \Big[\frac{\partial\mathbf{A}}{\partial\boldsymbol{\theta}}^\top \nabla_\mathbf{A} \mathcal{J} \Big]=-\nabla_{\boldsymbol{\theta}} \mathcal{L}_\text{BC}(\boldsymbol{\theta}, \mathbf{O}, \mathbf{A}^\text{target})$.
\end{theorem}

\begin{wrapfigure}{h}{0.6\textwidth}
    \vspace{-30pt}
    \begin{center}
    \includegraphics[width=0.6\textwidth]{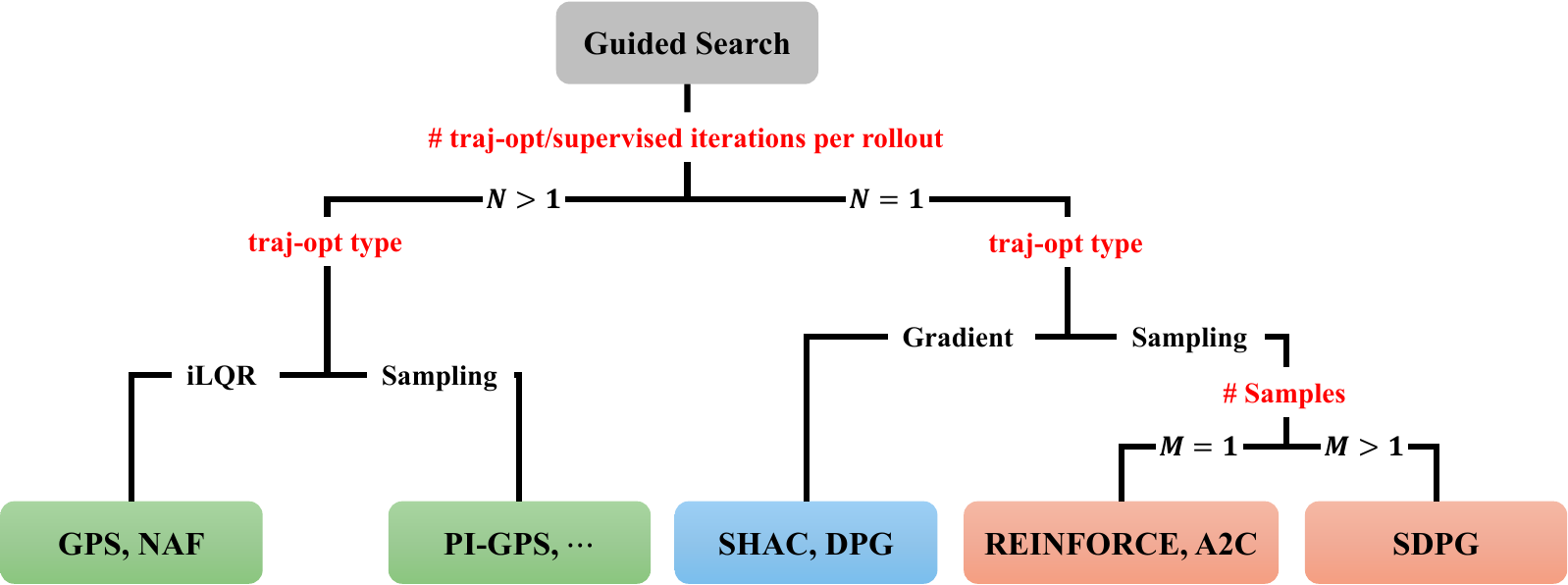}
    \end{center}
    \vspace{-10pt}
    \caption{\small{\textbf{Policy-based methods under the guided search view.} We show representative methods in each category, including GPS~\citep{pmlr-v28-levine13}, NAF~\citep{gu2016continuousdeepqlearningmodelbased}, PI-GPS~\citep{chebotar2018pathintegralguidedpolicy}, SHAC~\citep{xu2022acceleratedpolicylearningparallel}, DPG~\citep{you2025accelerating}, REINFORCE~\citep{williams1992simple}, A2C~\citep{mnih2016asynchronousmethodsdeepreinforcement}, and SDPG~(Ours). Based on the number of trajectory search and behavior cloning iterations per rollout batch, methods are organized into guided policy search methods (left) and policy gradient methods (right).}}
    \vspace{-10pt}
    \label{fig:guided_search}
\end{wrapfigure}

Theorem~\ref{theorem:connection_between_pg_and_gps} shows that policy gradients can be viewed as a special case of guided search: at each iteration, a batch of trajectories is rolled out, followed by a single step of trajectory optimization and a supervised policy update.
In contrast, classical guided policy search algorithms perform multiple iterations of trajectory optimization and supervised learning in the same rollout batch.
The SDPG we present above is a variant of policy gradient in which trajectory optimization is performed via stochastic sampling. 
We summarize these differences from the guided search perspective in Figure~\ref{fig:guided_search}.

An advantage of the guided search perspective is the explicit separation of hyperparameters for trajectory search and representation learning, enabling more efficient tuning. 
Moreover, the autoregressive loss we presented in~\eqref{eq: stochastic bc loss} and~\eqref{eq:behavior_cloning_loss} avoid likelihood ratio estimation, facilitating integration with high-capacity architectures such as diffusion models~\citep{yang2023policy,wagenmaker2025steering}, which we leave for future work.

\section{Practical algorithm design} \label{sec: pratical_design}
\begin{wrapfigure}{h}{0.5\textwidth}
    \vspace{-55pt}
    \begin{center}
    \includegraphics[width=0.5\textwidth]{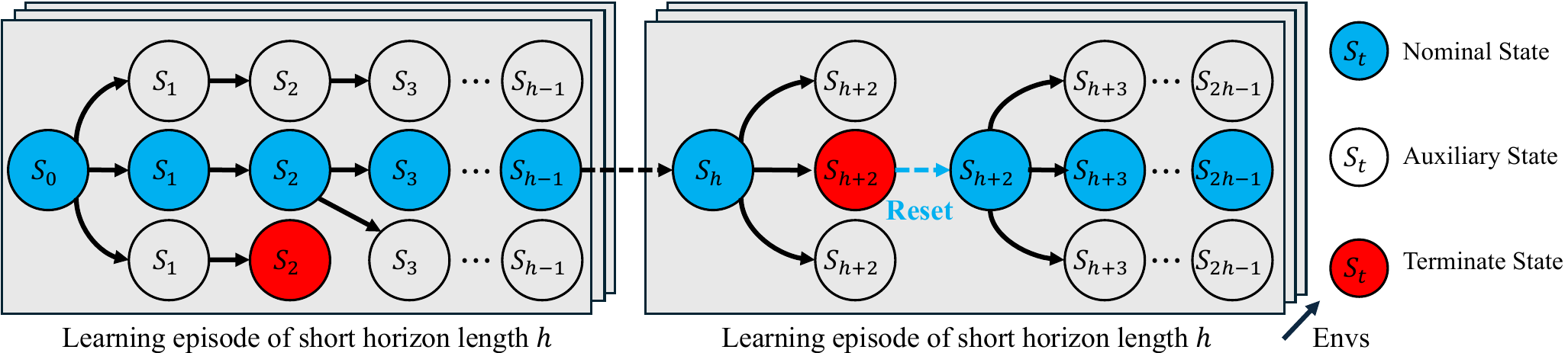}
    \end{center}
    \vspace{-10pt}
     \caption{\small{\textbf{Short-horizon rollout diagram.} At the start of each rollout segment, all auxiliary environments are reset to their corresponding nominal environment. Terminated auxiliary environments are reset to the nominal state, while termination of a nominal environment triggers a reset of all associated auxiliary environments.}}
    \vspace{-25pt}
    \label{fig: short_horizon rollouts}
\end{wrapfigure}
In this section, we present the key components that make SDPG practical; full algorithm, additional design choice and implementation details are deferred to Appendix~\ref{sec: additional impl details}.

\subsection{Actor–critic learning}

The actor-critic framework is a central component of policy gradient methods~\citep{sutton1998reinforcement, mnih2016asynchronousmethodsdeepreinforcement, schulman2017proximalpolicyoptimizationalgorithms}. 
In our work, we bootstrap return estimation by incorporating a value function at the final step:
\begin{equation}
    \mathcal{J}= \sum_{t=0}^H \gamma^t r_t + \gamma^{H+1} V_\phi(\mathbf{s}_{H+1}), \label{eq: short_horizon_return}
\end{equation}
where $H$ is horizon length, and the $V_\phi(\mathbf{s}_{H+1})$ is the value function learned by minimizing the supervised objective following standard TD$(\lambda)$ formulation~\citep{sutton1998reinforcement}:
\begin{equation}
    \mathcal{L}_\phi = \mathbb{E} [\|V_\phi(\mathbf{s}_t) - \mathrm{sg}(V_\text{target}(\mathbf{s}_t))\|]; \ V_\text{target}(\mathbf{s}_t) \coloneqq (1-\lambda) \big(\sum_{k=1}^{H-t-1} \lambda^k \mathcal{J}_t^k \big) + \lambda^{H-t-1}\mathcal{J}_t^{H-t}, \label{eq:critic loss}
\end{equation}
where $\mathcal{J}_t^k = (\sum_{l=0}^{k-1} \gamma^l r_{t+l}) + \gamma^k V_\phi(\mathbf{s}_{t+k})$ is the $k-$step return from time $t$. 
We define the critic on low-dimensional privileged states to accelerate learning, while the actor uses observations available on real sensors.
We also employ target networks for the critic to stabilize training, following common practice~\citep{mnih2013playingatarideepreinforcement,lillicrap2019continuouscontroldeepreinforcement,xu2022acceleratedpolicylearningparallel}.

\subsection{Short-horizon rollout}
Short-horizon rollouts divide long trajectories into shorter segments with updates after each segment, improving throughput and reducing variance.
We use $N$ nominal environments with batched rendering, each paired with $M$ state-only auxiliary environments to estimate $\nabla_{\boldsymbol{\theta}} \mathcal{V}_\text{SDPG}$~\eqref{eq: sdpg derivation}.
Reset and termination are handled separately between nominal and auxiliary environments.
Figure~\ref{fig: short_horizon rollouts} illustrates the procedure; full details are provided in Appendix~\ref{sec: rollout proce}.

\subsection{Learning exploration factor}

We find that dynamically adapting the exploration factor $\delta$ in Equation~\eqref{eq:smoother_action_improved_direction} improves convergence. 
Similar to the mean-action update in~\eqref{eq:smoother_action_improved_direction} and~\eqref{eq: stochastic bc loss}, we update the exploration factor $\tilde{\delta} \coloneqq \log \delta$ via (see Appendix~\ref{sec: derivation of exploration} for derivation): 
\begin{equation}
    \mathcal{L}_{\tilde{\delta}} \coloneqq \mathbb{E} \Big[ \|\tilde{\delta} - \mathrm{sg}(\tilde{\delta}^\text{target})\|_2^2\Big], \ \tilde{\delta}^\text{target} \coloneq \tilde{\delta} + \frac{\mathbf{I}}{T}  \mathbb{E}_{\mathcal{E} \sim \mathcal{N}(\mathbf{0}, \mathbf{I})} \Big[ \frac{\delta}{\sigma}\big (\mathcal{J}(\mathbf{A} + \delta \mathcal{E}) - \mathcal{J}(\mathbf{A})\big) (\mathcal{E}^2 - 1) \Big]. \label{eq: exploration update strategy}
\end{equation}
Here, we keep the exploration factor in the same dimensional space as the action, i.e. $\dim(\tilde{ \delta}) = \dim(\mathcal{A})$ and $\mathbf{I}$ is a projection matrix.
We also incorporate entropy terms to prevent premature convergence~(See Appendix~\ref{sec: entropy} for details).

\section{Experiments} \label{sec: exp}
\begin{figure}[t]
    \vspace{-15pt}
    \begin{center}
    \includegraphics[width=\textwidth]{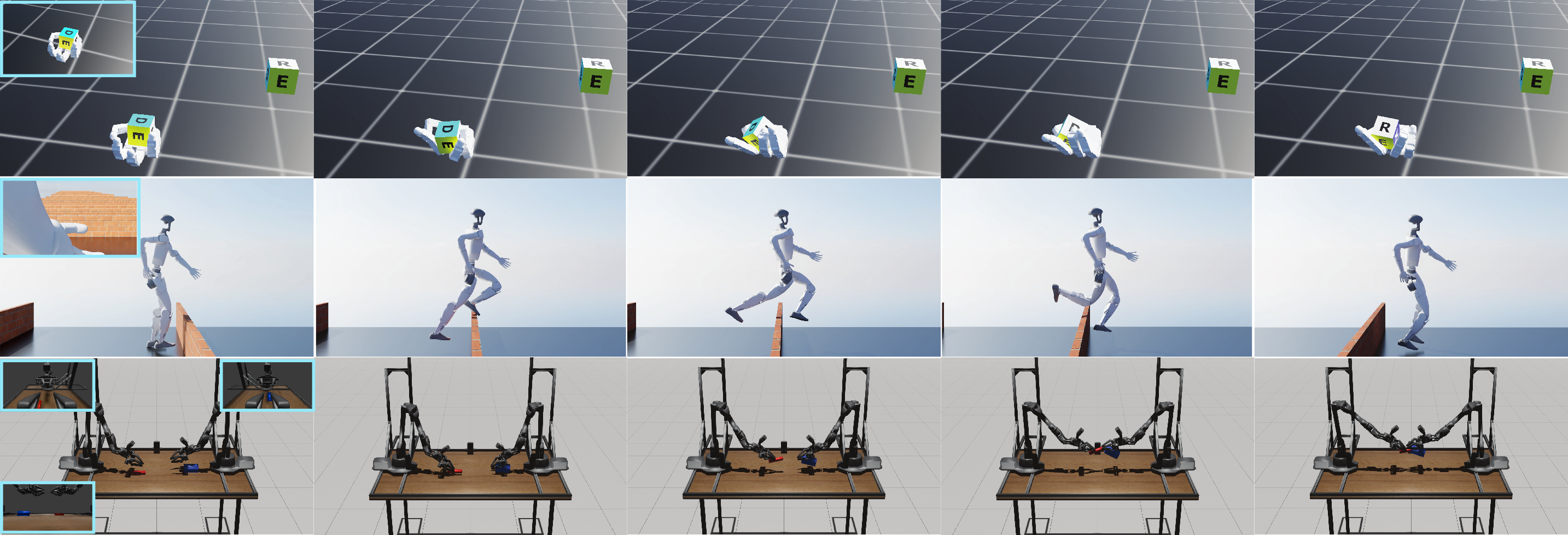}
    \end{center}
    \vspace{-10pt}
    \caption{\small{\textbf{Illustration of learned policies:} a sequence of frames from trajectories learned by our method. Examples of camera views are presented in the first frame.}~\textbf{Top}: Allegro hand reorienting a cube.~\textbf{Mid}: G1 hurdling.~\textbf{Bottom}: Aloha performs insertion.}
    \vspace{-15pt}
    \label{fig:sim_traj}
\end{figure}

\begin{wraptable}{r}{0.50\textwidth}
  \vspace{-67pt}
  \footnotesize
  \renewcommand{\arraystretch}{0.85}
  \setlength{\tabcolsep}{3pt}
  \vspace{-0pt}
  \caption{\footnotesize{Memory usage (GB). Except PPO, all methods are trained with $64$ batch rendered envs.}}
  \label{tab: memory usage}
  \vspace{5pt}
  \centering
  \begin{threeparttable}
  \begin{tabular}{ccccc}
    \toprule
     &Hopper &Walker &Ant &Humanoid \\
    \midrule
    \rowcolor{blue!10}
    Ours &10.2 &10.3 &10.3 &10.5\\
    \midrule
    \rowcolor{red!10}
    PPO\tnote{$\dagger$}  & 48$^{\dagger}$ & 48$^{\dagger}$ & 49$^{\dagger}$& 50$^{\dagger}$ \\
    \midrule
    DrQv2 &10.6 &8.2 &10.5 &11.6\\
    \midrule
    DreamerV3 &10.8 &10.8 &10.8 &10.9 \\
    \midrule
    Distillation &10.6 &10.6 &10.3 &10.7\\
    \bottomrule
  \end{tabular}
  \begin{tablenotes}
    \footnotesize
    \item[$\dagger$] PPO memory is estimated with 4096 environments and state-based hyperparameters.
  \end{tablenotes}
  \end{threeparttable}
  \vspace{-15pt}
\end{wraptable}

To validate the performance of our method, we first compare it on standard benchmarks against popular visual policy learning approaches. 
To further demonstrate its effectiveness for robotics, we apply our method to several challenging egocentric tasks spanning both locomotion and manipulation. 
Finally, we evaluate its sim-to-real transfer performance on hardware.
All environment and training details, including rewards and hyperparameters, are provided in the Appendix~\ref{sec: add exp setting}.

\subsection{Benchmarks}
\paragraph{Settings}
We use Visual MuJoCo as our testbed, where the algorithm must learn to control the robot from third-person RGB images. 
The environments are reimplemented in Genesis simulation~\citep{Genesis}, which provides fast physics and batch rendering, and we replicate the rewards and other environment settings from~\citet{you2025accelerating}.
All baselines are evaluated on a single NVIDIA RTX 4080 GPU.

\paragraph{Baselines}
\begin{figure}[t]
    \vspace{-20pt}
    \begin{center}
    \includegraphics[width=\textwidth]{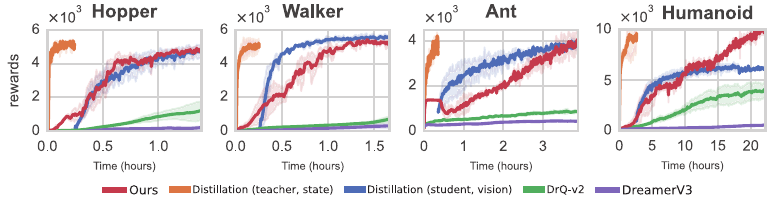}
    \end{center}
    \vspace{-10pt}
    \caption{\small{\textbf{Mujoco Benchmark:} Despite being end-to-end, our method matches distillation in training speed, is significantly faster than DrQ-v2 and DreamerV3, and achieves higher final rewards on humanoid tasks.}}
    \vspace{-10pt}
    \label{fig:mujoco_benchmark}
\end{figure}

We compare several popular methods for visual policy learning.
\textit{DrQv2}~\citep{yarats2021masteringvisualcontinuouscontrol}: A strong off-policy method for visual-motor control. 
We augment the original implementation with parallel simulation and code optimizations for fair wall-clock comparison, reducing runtime by ~2× on the same machine.
\textit{DreamerV3}~\citep{hafner2024masteringdiversedomainsworld}: A model-based method capable of training visual policies. We apply similar code optimizations; however, the main bottleneck is the world model training, and code the improvements provide minimal speedup in wall-clock time.
\textit{Teacher-Student Distillation}: A widely used method for visuomotor control.
We follow the improvement in~\citet{mu2024preferstatetovisualdaggervisual} to ensure both wall-clock efficiency and strong overall performance.
The teacher policy are trained with RL-games~\citep{rl-games2021} and RSL-RL~\citep{schwarke2025rslrl} PPO.

\paragraph{Results}
\begin{wrapfigure}{h}{0.3\textwidth}
    \vspace{-45pt}
    \begin{center}
    \includegraphics[width=0.3\textwidth]{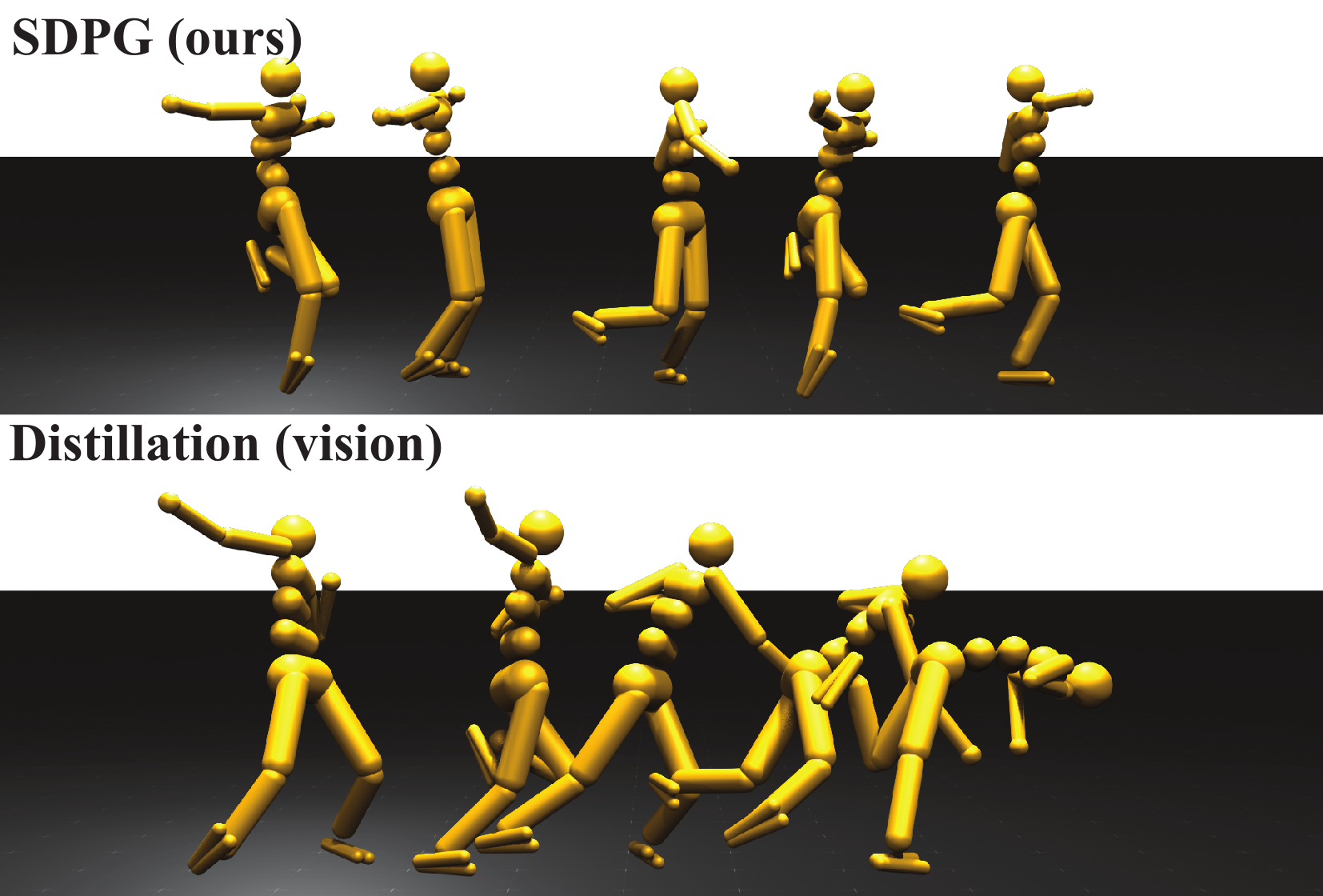}
    \end{center}
    \vspace{-10pt}
    \caption{\small {On humanoid tasks, distillation-based visual policies plateau at suboptimal performance and fail in certain states, whereas ours remains stable and runs continuously.}}
    \vspace{-12pt}
    \label{fig:humanoid_traj}
\end{wrapfigure}
Figure~\ref{fig:mujoco_benchmark} shows the training curves. Our method achieves the highest rewards, matching state-based performance across all tasks. 
It is competitive with teacher–student distillation in training time and significantly faster than other end-to-end visual RL methods. 
While we do not directly compare to~\citet{you2025accelerating} due to the lack of rigid-body differentiability in Genesis at the time of this work, the reported training times are comparable, owing to the elimination of the backward pass and the faster simulation speed of Genesis relative to Dflex~\citep{xu2022acceleratedpolicylearningparallel}. Despite being on-policy, our method has memory usage comparable to off-policy and model-based approaches, enabled by accurate gradient estimation with a small number of batched environments (e.g. 64 vs. 4096 in PPO), allowing training on a single NVIDIA RTX 4080 GPU.

\begin{wrapfigure}{h}{0.42\textwidth}
    \vspace{-15pt}
    \begin{center}
    \includegraphics[width=0.42\textwidth]{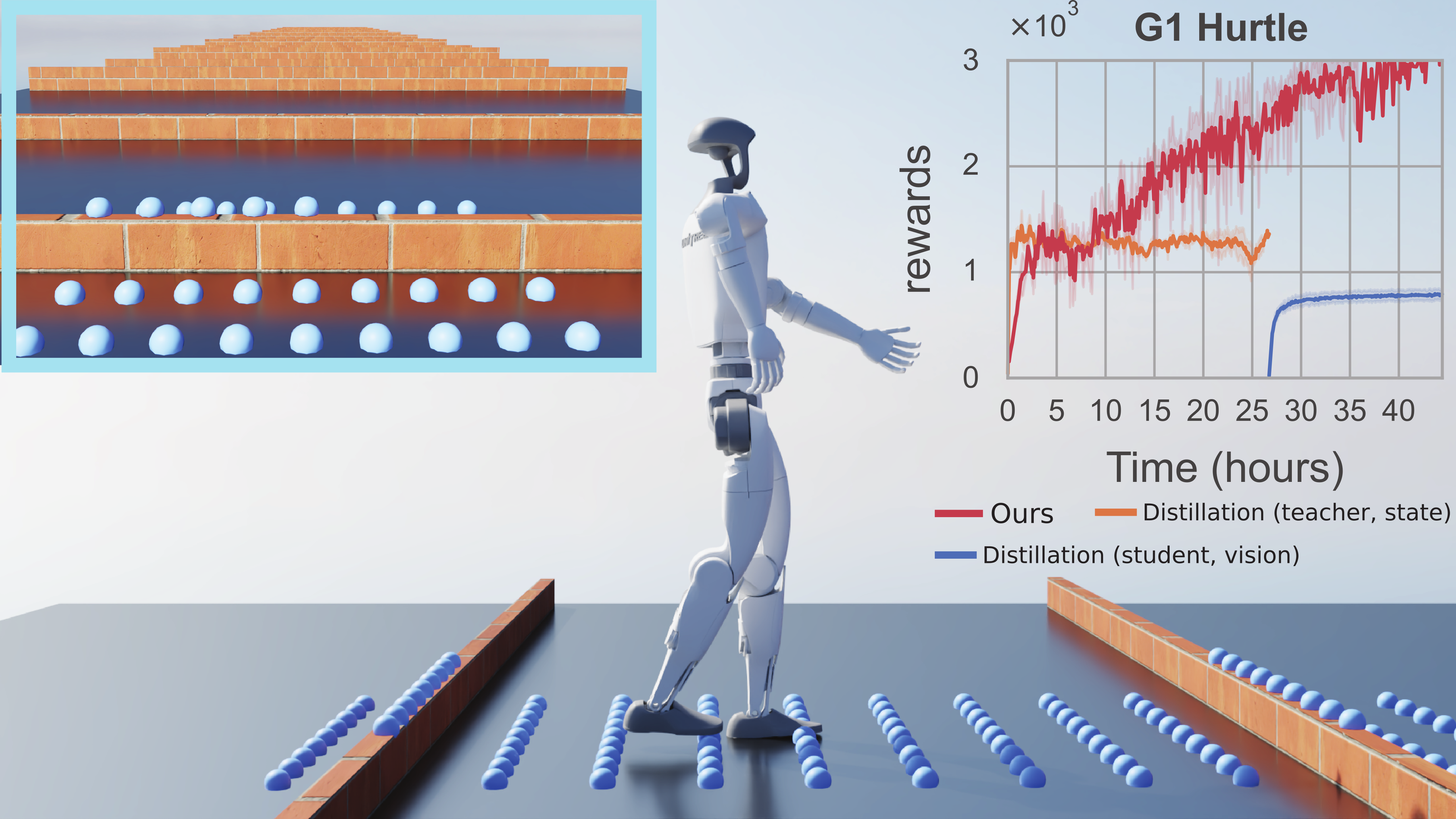}
    \end{center}
    \vspace{-10pt}
    \caption{\small{\textbf{G1 hurdle example:} The left panel illustrates the effective observation range of height-map (blue dots) and RGB inputs, while the right panel shows the training curves. Training a teacher policy with local height maps can be more difficult than learning directly from visual observations end-to-end.}}
    \vspace{-20pt}
    \label{fig:g1_hurtle}
\end{wrapfigure}
\textbf{When to prefer visual RL over teacher–student distillation}
For moderately difficult tasks, teacher--student distillation can match visual RL performance, but the gap widens on more challenging problems~\citep{rudin2025parkourwildlearninggeneral, singh2025endtoendrlimprovesdexterous}. In humanoid tasks, we observe that students frequently visit states underrepresented in the teacher data~(unstable or falling down configuration), where supervision becomes less effective, consistent with prior findings~\citet{you2025accelerating, mu2024preferstatetovisualdaggervisual}. 
This also makes distillation harder to tune, as improving early training speed often sacrifices final performance. 
In contrast, visual RL learns directly from its own experience and is more robust in these regimes.

We also find that designing effective low-dimensional observations for teacher training can be nontrivial. For example, in the G1 hurdle task, we use height maps commonly adopted in locomotion literature~\citep{rudin2022learningwalkminutesusing} to represent terrain. However, height maps only capture local geometry and have limited effective range, which can make teacher training difficult. 
In contrast, RGB observations naturally provide richer long-range information. Moreover, large pretrained visual encoders~\citep{siméoni2025dinov3, radford2021learningtransferablevisualmodels} are now widely available, whereas pretrained encoders for low-dimensional modalities such as meshes or point clouds are far less common, making direct visual policy learning increasingly attractive.

\begin{figure}[t]
    \vspace{-15pt}
    \begin{center}
    \includegraphics[width=\textwidth]{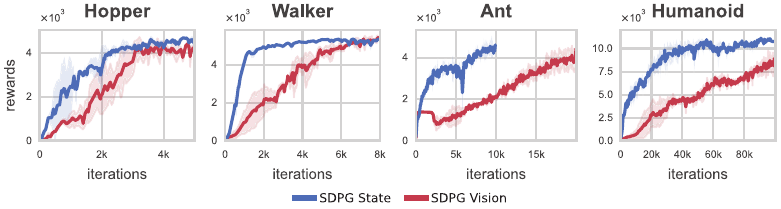}
    \end{center}
    \vspace{-10pt}
    \caption{\small{\textbf{Comparison between state and visual inputs.} The performance is consistent across modalities, while visual tasks typically require more iterations to converge.}}
    \vspace{-15pt}
    \label{fig:state_vis_comparison}
\end{figure}

\textbf{Learning state vs vision inputs}
The guided search view separates trajectory search from representation learning, allowing hyperparameters to be tuned independently. 
In practice, we tune search-related hyperparameters using low-dimensional state inputs for faster iteration, then transfer them directly to visual settings by adding a CNN encoder.
Figure~\ref{fig:state_vis_comparison} show the training curve with state and visual observations under identical hyperparameters (except an additional CNN encoder). 
Final rewards are similar across modalities, while visual inputs require 2--4$\times$ more iterations to converge.

\subsection{Ego-centric suite}

To facilitate further research, we provide a suite of challenging robotics tasks spanning dexterous manipulation and locomotion. 
The robot receives both egocentric images and proprioceptive measurements, mimicking real hardware sensors. 
Figure~\ref{fig:preview} provides a preview of the tasks, while Figure~\ref{fig:sim_traj} shows example trajectories generated by the learned policy.
Details on each task are provided in Appendix~\ref{sec: ego centric tasks}, while Appendix~\ref{sec: eco-centric training} provides example training curves from our method.

\subsection{Sim-to-real experiment}

\begin{figure}[h]
    \vspace{-5pt}
    \begin{center}
    \includegraphics[width=\textwidth]{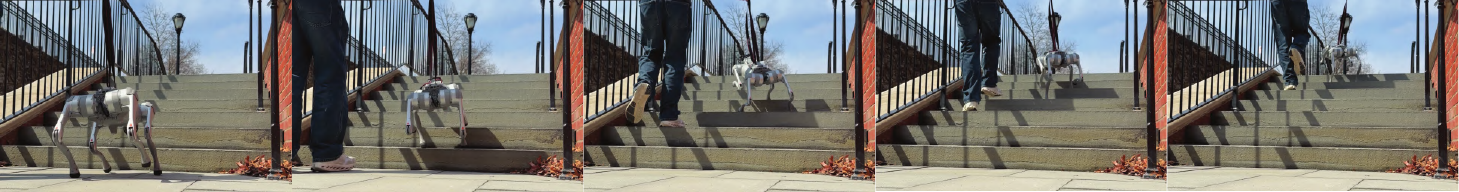}
    \end{center}
    \vspace{-10pt}
    \caption{\small We train Go2 with depth camera on diverse terrains and transfer zero-shot to the real stair.}
    \vspace{-15pt}
    \label{fig:go2_stair_real}
\end{figure}

Finally, we validate our method on Go2 hardware, where the robot uses an ego-centric RealSense camera to perceive the environment~(depth) and navigate challenging terrains, including uneven surfaces, boxes, and stairs. 
The policy is trained entirely in simulation within 2 hours and transferred to the real world via zero-shot sim-to-real transfer.
Figure~\ref{fig:go2_stair_real} demonstrate the Go2 walk through an upstair, while experiments on other types of terrain are available in Appendix~\ref{sec: add real exp}. 

\section{Limitation and future work} \label{sec: lim_and_future_work}

While our method is computationally efficient in its design, its overall cost remains dominated by physical simulation, as rollouts are discarded after each iteration.
In this sense, our method is closer to early A2C-style methods, whereas approaches such as PPO reuse samples for multiple updates.
We also observe some sensitivity to random seeds and occasional training instability. 
Incorporating techniques from PPO and related first-order methods is a promising direction to further improve both training efficiency and stability.
\section{Conclusion} \label{sec: conclusion}

We present SDPG, a computationally efficient policy gradient method for visual RL. 
Through simple random search, SDPG enables stable end-to-end training with modest computational resources.
We evaluate SDPG in a diverse set of robotic tasks, demonstrating its effectiveness across challenging visual control problems. 
Its strong wall-clock and memory efficiency make end-to-end visual RL more accessible and scalable to complex settings.
We also provide a new perspective on policy gradient methods, highlighting their connection to classical trajectory optimization and showing how this viewpoint can inform practical design and hyperparameter choices.

\clearpage
\acknowledgments{}


\bibliography{reference}  
\appendix
\clearpage
\section{Related work}\label{sec: related work}

\subsection{Visual-RL}

Early works such as~\cite{mnih2013playingatarideepreinforcement, mnih2015human} demonstrated that RL can operate directly on raw pixel observations, achieving human-level performance on discrete control tasks such as Atari games and setting the foundation for visual-RL.

For continuous control, a line of works including~\citep{pinto2017asymmetricactorcriticimagebased, yarats2021masteringvisualcontinuouscontrol} extended off-policy actor-critic methods such as DDPG~\citep{lillicrap2019continuouscontroldeepreinforcement} and SAC~\citep{srinivas2020curlcontrastiveunsupervisedrepresentations, yarats2021image} to complex tasks. 
Model-based approaches provide another promising direction by learning predictive dynamics models for planning or policy optimization. Representative methods include TP-MPC~\citep{hansen2022temporaldifferencelearningmodel, hansen2024tdmpc2}, Dreamer~\citep{hafner2020dreamcontrollearningbehaviors,hafner2022masteringataridiscreteworld,hafner2024masteringdiversedomainsworld}, and more recent JEPA-style models~\citep{assran2025vjepa2selfsupervisedvideo,garrido2024learningleveragingworldmodels,bardes2024revisiting} series. 
However, both off-policy and model-based method are slow to train and sensitive to hyperparameters as they need to update the complex internal representation.

In contrast, on-policy methods such as PPO~\citep{zakka2025mujocoplayground,tao2025maniskill3gpuparallelizedrobotics, singh2025endtoendrlimprovesdexterous} offer greater training efficiency and stability, and have become a popular choice for visual control tasks.
However, their main limitation lies in memory consumption due to large-scale parallel simulation and rendering.

First-order methods~\citep{you2025accelerating, Zhang_2025,pan2026learningflyrapidpolicy} based on differentiable simulation offer another alternative for visual-RL. However, they often struggle in high-dimensional and noisy optimization landscapes and can be less flexible due to limited support in simulation ecosystems and toolchains.

In contrast to these approaches, our method is both computation- and memory-efficient and is designed to integrate easily into existing robotics software systems.

\subsection{Evolutionary strategy}
Evolutionary strategy~\citep{huning1976evolutionsstrategie} learns by iteratively perturbing parameters with stochastic noise and selecting updates that improve performance.
\citet{salimans2017evolutionstrategiesscalablealternative, mania2018simplerandomsearchprovides} show that directly applying evolutionary strategies in parameter space can achieve performance competitive with reinforcement learning methods.
Evolutionary strategies are also widely used in modern control, particularly as optimization solvers within model predictive control (MPC) frameworks~\citep{howell2022predictivesamplingrealtimebehaviour,williams2016aggressive}, commonly referred to as sampling-based MPC.

In this work, we apply evolutionary strategies in the action space, similar in spirit to RL and sampling-based MPC, and show that with appropriate algorithmic and systems design, they can provide an efficient approach for vision-based control.
\section{Additional proof}\label{sec:additional proof}

\subsection{Derivation of exploration factor update rules} \label{sec: derivation of exploration}
We derive the update rule for the exploration factor in~\eqref{eq: exploration update strategy} from the policy gradient objective under a Gaussian policy:
\begin{equation}
    \mathcal{L}_\text{PG} = \mathbb{E} \Big [ (\mathcal{J}-\mathcal{B})\sum_{t=0}^T \log \pi(\mathbf{u_t}|\mathbf{o_t})\Big] = \mathbb{E} \Big\{\frac{\mathcal{B}-\mathcal{J}}{2} \sum_{t=0}^T \big[(\mathbf{u}_t-\mathbf{a}_t)^\top\boldsymbol{\Sigma}^{-1}(\mathbf{u}_t-\mathbf{a}_t) + \log\det \boldsymbol{\Sigma}\big]\Big\} + \text{const},
\end{equation}
where $\mathbf{a}_t = \boldsymbol{\mu}(\mathbf{o}_t)$, $\mathbf{u}_t = \mathbf{a}_t + \exp(\boldsymbol{\tilde{\delta}})\boldsymbol{\epsilon}_t$, and $\boldsymbol{\Sigma} = \mathrm{diag}(\exp(\boldsymbol{\tilde{\delta}})^2)$.

This yields the gradients
\begin{equation}
    \nabla_{\mathbf{A}}\mathcal{L}_\text{PG} = \frac{(\mathcal{J - \mathcal{B}}) \boldsymbol{\epsilon}_t}{\delta}; \ \nabla_{\boldsymbol{\tilde{\delta}}} \mathcal{L}_\text{PG} = (\mathcal{J}-\mathcal{B})( {\mathcal{E}^2 -1)}.
\end{equation}

Using $\mathcal{J}(\mathbf{A})$ as the baseline, we construct the target exploration update as:
\begin{equation}
    \frac{\delta}{\sigma}\big (\mathcal{J}(\mathbf{A} + \delta \mathcal{E}) - \mathcal{J}(\mathbf{A})\big) (\mathcal{E}^2 - 1)
\end{equation}
Here, division by $\sigma$ normalizes the return, while scaling by $\delta$ preserves the relative magnitude between the mean and variance updates.
In practice, very small $\delta$ leads to large gradients $\propto \frac{1}{\delta}$, causing instability in the mean update. 
Thus, we scale both mean and exploration updates by $\delta$.

\subsection{Proof of Theorem~\ref{thm:stochastic_smoothing}} \label{sec: proof of thm stochastic}

We have 

\begin{equation}
\begin{aligned}
    \nabla_\mathbf{A} \mathcal{J}_\delta = &\nabla_\mathbf{A} \int  \mathcal{J}(\mathbf{A} + \delta \mathcal{E}) p(\mathcal{E}) d \mathcal{E}\\
    = & \nabla_\mathbf{A} \int \frac{1}{\delta^d}\mathcal{J}(\mathbf{U}) p(\mathbf{\frac{U -A}{\delta}}) d\mathbf{U} \\
    = & \int \frac{1}{\delta^d}\mathcal{J}(\mathbf{U}) \nabla_\mathbf{A} p(\frac{\mathbf{U}-A}{\delta}) d\mathbf{U} \\
    = &\int \frac{1}{\delta^d}\mathcal{J}(\mathbf{U}) p(\frac{\mathbf{U} -\mathbf{A}}{\delta}) \nabla_\mathbf{A} \log p(\frac{\mathbf{U} -\mathbf{A}}{\delta}) d\mathbf{U} \\
    =& \int \frac{1}{\delta^d}\mathcal{J}(\mathbf{U}) \rho(\frac{\mathbf{U}-\mathbf{A}}{\delta}) \frac{\mathbf{U}-\mathbf{A}}{\delta^2} d\mathbf{U} \\
    =& \mathbb{E}\Big[\frac{\mathcal{J}(\mathbf{A} + \delta \mathcal{E})}{\delta} \mathcal{E}\Big]\\
    =& \mathbb{E}\Big[\frac{\mathcal{J}(\mathbf{A} + \delta \mathcal{E}) - \mathcal{J}(\mathbf{A})}{\delta} \mathcal{E}\Big]\\
    =&\frac{\delta}{\sigma} \nabla \mathcal{J}_\text{smooth}(\mathbf{A})
\end{aligned}
\end{equation}

\subsection{Proof of Proposition~\ref{prop: update rule and supervised trick}} \label{sec: proof of prop}
We have 
\begin{equation}
    \nabla_{\mathbf{x}_k}\mathcal{L} = \frac{d \mathbf{y}_k}{d\mathbf{x}_k}^\top (\mathbf{y}_k -\mathbf{y}_k^\text{target}) =-\frac{d\mathbf{y}_k}{d \mathbf{x}_k}^\top \mathbf{d}(\mathbf{y}_k).
\end{equation}
\subsection{Proof of Theorem~\ref{theorem:connection_between_pg_and_gps}} \label{sec: proof of pg and gps}
We can rewrite the behavior cloning objective in~\eqref{eq:behavior_cloning_loss} as
\begin{equation}
    \mathcal{L}_\text{BC}(\boldsymbol{\theta}, \mathbf{O}, \mathbf{A}^\text{target}) = \frac{1}{2 \alpha} \mathbb{E}_\tau [\|\mathbf{A} - \mathbf{A}^\text{target}\|^2_2].
\end{equation}
Via Proposition~\ref{prop: update rule and supervised trick},
we then have
\begin{equation}
    \mathcal{L}_\text{BC}(\boldsymbol{\theta}, \mathbf{O}, \mathbf{A}^\text{target}) = -\frac{1}{2 \alpha} \mathbb{E}_\tau \Big[\frac{\partial \mathbf{A}}{\partial \boldsymbol{\theta}}^\top \alpha \nabla_\mathbf{A}\mathcal{J}\Big] = -\nabla_{\boldsymbol{\theta}} \mathcal{V}_\text{decoupled}
\end{equation}
\section{Additional implementation details}\label{sec: additional impl details}
\subsection{Rollout and update procedure}\label{sec: rollout proce}
We initialize $N$ nominal environments with batched rendering and $M$ state-only auxiliary environments for each nominal. 
In total, there are $N(M+1)$ physical environments, but only $N$ require batch rendering.
At the start of each rollout segment, auxiliary environments are reset to the same internal states (e.g., joint positions and velocities) as their corresponding nominal environments. 
In each step, nominal environments generate rendered observations to compute actions, which are then perturbed and applied in the auxiliary environments.
During simulation, the terminated auxiliary environments are reset to the corresponding nominal state. 
If a nominal environment terminates, it is reset along with all its associated auxiliary environments. 
After each rollout segment, we compute returns according to Equation~\eqref{eq: short_horizon_return}, and update the actor via one-step gradient descent on the supervised loss~\eqref{eq: stochastic bc loss} and the critic via~\eqref{eq:critic loss}. 
The next rollout segment continues from the final states of the previous one.

\subsection{Entropy Terms}\label{sec: entropy}

Here, we describe how the entropy term is incorporated.

\paragraph{Entropy as regularization}

We find that adding an entropy term to the actor $\pi$ improves exploration and helps prevent premature convergence to suboptimal policies. This idea was introduced in~\citet{williams1991function} and is widely used in actor-critic methods such as A2C~\citep{mnih2016asynchronousmethodsdeepreinforcement}.

For a diagonal Gaussian policy, the Shannon entropy is
\begin{equation}
    H = \sum_{i=1}^d \left(\log \delta_i + \frac{1}{2} + \frac{1}{2}\log(2\pi)\right),
\end{equation}
where $d$ is the action dimension. Let $\boldsymbol{\tilde{\delta}} \coloneqq \log \boldsymbol{\delta}$. 
Then $\nabla_{\boldsymbol{\tilde{\delta}}} H = \mathbf{1}$.
We incorporate entropy regularization by modifying the exploration objective in~\eqref{eq: exploration update strategy} as
\begin{equation}
    \mathcal{L}_{\boldsymbol{\tilde{\delta}}}
    =
    \mathbb{E}\Big[
    \|\boldsymbol{\tilde{\delta}}-\mathrm{sg}(\boldsymbol{\tilde{\delta}}^{\text{target}})\|_2^2
    -
    \alpha \frac{1}{d}\sum_{i=1}^{d}\tilde{\delta}_i
    \Big],
\end{equation}
where $\alpha$ controls the strength of entropy regularization.
The form is analogous to weight decay in supervised learning and we therefore refer to it as entropy regularization.

\paragraph{Soft critic}

To further encourage exploration, entropy can also be incorporated into critic learning, which is a central idea behind methods such as SAC~\citep{haarnoja2018softactorcriticoffpolicymaximum,haarnoja2019softactorcriticalgorithmsapplications}.

We adopt a soft actor-critic style target by augmenting $\mathcal{J}_t^k$ in~\eqref{eq:critic loss} as
\begin{equation}
\mathcal{J}_t^k =
\Big[
\sum_{l=0}^{k-1}\gamma^l
\Big(
r_{t+l}
+
\frac{\alpha}{d}\sum_{i=1}^{d}(\tilde{\delta}_i-\tilde{\delta}_{\text{target}})
\Big)
\Big]
+
\gamma^k V_\phi(\mathbf{s}_{t+k}),
\end{equation}
where $\frac{\alpha}{d}\sum_{i=1}^{d}(\tilde{\delta}_i-\tilde{\delta}_{\text{target}})$ is the added entropy reward.

This reward is a linear transformation of Shannon entropy that is normalized by action dimension and shifted to be zero at the target exploration level. It is also more interpretable, as it directly reflects deviation from the desired exploration scale (e.g., action-joint variance).

\paragraph{Temperature auto-tuning}

We further incorporate an automatic temperature tuning mechanism similar to~\citet{haarnoja2019softactorcriticalgorithmsapplications}.
The temperature is learned by optimizing the following objective
\begin{equation}
    \mathcal{L}(\tilde{\alpha}) = \exp({\tilde{\alpha}}) \sum_{i=1}^d (\tilde{\delta}_i-\tilde{\delta}_{\text{target}}); \ \tilde{\alpha}\coloneqq \log \alpha,
\end{equation}
where $\alpha$ is parameterized in log-space to ensure positivity. 
We also replace the entropy target with a target standard deviation, making the objective more interpretable and consistent across tasks.

\paragraph{Recommendations}
We test both state-dependent and state-independent exploration parameterizations.
We experiment with applying entropy in both the actor and critic, along with automatic temperature tuning.
Although the decision choice are task specific and also related to code implementation, we find, in general, that a state-independent exploration factor combined with entropy regularization in the actor performs well across tasks and is easier to tune.
This setting is similar to early implementations of A3C~\citep{mnih2016asynchronousmethodsdeepreinforcement} and PPO~\citep{schulman2017proximalpolicyoptimizationalgorithms}.

\subsection{Causality and eligibility trace}

\begin{algorithm}[h]
\caption{SDPG: Stochastic Decoupled Policy Gradient}
\label{alg: sdpg}
\begin{algorithmic}[1]
\State \textbf{Input:} actor $\pi_{\boldsymbol{\theta}}$, critic $V_\phi$, target critic $V_{\phi'}$, exploration factor $\boldsymbol{\tilde{\delta}}$, log-temperature $\tilde{\alpha}$
\State \textbf{Hyperparameters:} nominal envs $N$, auxiliary envs per nominal $M$, segment length $H$, discount $\gamma$, TD($\lambda$) factor $\lambda$, target-critic coeff.\ $\rho$, target std $\delta^{\text{target}}$, critic iters $K_c$, mini-batch size $B$, learning rates $\eta_{\boldsymbol{\theta}}, \eta_\phi, \eta_{\tilde{\delta}}, \eta_{\tilde{\alpha}}$
\State Initialize $N(M+1)$ parallel environments; copy each nominal initial state to its $M$ auxiliaries
\For{$\text{epoch}=1,2,\dots$}
    \State \textit{// Rollout segment of length $H$}
    \For{$t=0,\dots,H-1$}
        \State Get observations $\mathbf{o}_t^{(n)}$ from the $N$ nominal envs
        \State Compute $\mathbf{a}_t^{(n)} \gets \boldsymbol{\mu}_{\boldsymbol{\theta}}(\mathbf{o}_t^{(n)})$ and $\boldsymbol{\delta}\gets\exp(\boldsymbol{\tilde{\delta}})$
        \State Sample $\boldsymbol{\epsilon}_t^{(n,j)}\sim\mathcal{N}(\mathbf{0},\mathbf{I})$ for $j=1,\dots,M$; set $\boldsymbol{\epsilon}_t^{(n,0)}\gets\mathbf{0}$
        \State Form actions $\mathbf{u}_t^{(n,j)} \gets \mathbf{a}_t^{(n)} + \boldsymbol{\delta}\odot\boldsymbol{\epsilon}_t^{(n,j)}$ for all $(n,j)$
        \State Step every env with $\tanh(\mathbf{u}_t^{(n,j)})$; collect rewards $r_t^{(n,j)}$ and next states $\mathbf{s}_{t+1}^{(n,j)}$
        \State Query target critic $V_{\phi'}(\mathbf{s}_{t+1}^{(n,j)})$ for all non-terminated envs (zero otherwise)
        \State \textbf{Reset handling:} if a nominal env $n$ terminates, reset $n$ and copy its new state to all its auxiliaries; if only an auxiliary $(n,j)$ terminates, copy the current nominal state into it
    \EndFor
    \State \textit{// Causal TD($\lambda$) return-to-go} (Sec.~\ref{sec: rollout proce}, Eq.~\eqref{eq:critic loss})
    \For{$t=H-1,\dots,0$ and every env $(n,j)$}
        \State Compute the TD($\lambda$) return $\mathcal{J}_t^{(n,j)}$ from $\{r_{t:H-1}^{(n,j)}, V_{\phi'}\}$, restarting at termination boundaries
    \EndFor
    \State \textit{// Actor target via SDPG} (Eqs.~\eqref{eq:smoother_action_improved_direction},~\eqref{eq: exploration update strategy})
    \State $\Delta\mathcal{J}_t^{(n,j)} \gets \mathcal{J}_t^{(n,j)} - \mathcal{J}_t^{(n,0)}$ \Comment{causal $\Delta\mathcal{J}$ relative to nominal}
    \State Optionally normalize $\Delta\mathcal{J}_t^{(n,j)}$ by its per-$(n,t)$ standard deviation across $j$
    \State Mean ascent: $\mathbf{g}_{\boldsymbol{\mu},t}^{(n)} \gets \frac{1}{M+1}\sum_{j=0}^{M}\Delta\mathcal{J}_t^{(n,j)}\,\boldsymbol{\epsilon}_t^{(n,j)}$
    \State Exploration ascent: $\mathbf{g}_{\tilde{\delta},t}^{(n)} \gets \frac{1}{M+1}\sum_{j=0}^{M}\Delta\mathcal{J}_t^{(n,j)}\,\big(\boldsymbol{\epsilon}_t^{(n,j)}\odot\boldsymbol{\epsilon}_t^{(n,j)}-\mathbf{1}\big)\odot\boldsymbol{\delta}$
    \State Actor targets: $\mathbf{a}_t^{(n),\text{target}} \gets \mathrm{sg}\!\big(\mathbf{a}_t^{(n)} + \mathbf{g}_{\boldsymbol{\mu},t}^{(n)}\big)$; \ $\boldsymbol{\tilde{\delta}}^{\text{target}} \gets \mathrm{sg}\!\big(\boldsymbol{\tilde{\delta}} + \tfrac{1}{NH}\sum_{n,t}\mathbf{g}_{\tilde{\delta},t}^{(n)}\big)$
    \State \textit{// Actor + exploration-factor update} (Eqs.~\eqref{eq: stochastic bc loss},~\eqref{eq: exploration update strategy})
    \State $\mathcal{L}_{\boldsymbol{\theta},\boldsymbol{\tilde{\delta}}} \gets \tfrac{1}{NH}\sum_{n,t}\big\|\boldsymbol{\mu}_{\boldsymbol{\theta}}(\mathbf{o}_t^{(n)}) - \mathbf{a}_t^{(n),\text{target}}\big\|_2^2 \;+\; \big\|\boldsymbol{\tilde{\delta}} - \boldsymbol{\tilde{\delta}}^{\text{target}}\big\|_2^2 \;-\; \alpha\,\tfrac{1}{d}\sum_{i=1}^{d} \tilde{\delta}_i$
    \State $\boldsymbol{\theta} \gets \boldsymbol{\theta} - \eta_{\boldsymbol{\theta}}\nabla_{\boldsymbol{\theta}}\mathcal{L}_{\boldsymbol{\theta},\boldsymbol{\tilde{\delta}}}$; \quad $\boldsymbol{\tilde{\delta}} \gets \boldsymbol{\tilde{\delta}} - \eta_{\tilde{\delta}}\nabla_{\boldsymbol{\tilde{\delta}}}\mathcal{L}_{\boldsymbol{\theta},\boldsymbol{\tilde{\delta}}}$
    \State \textit{// Temperature auto-tune} (Sec.~\ref{sec: entropy})
    \State $\mathcal{L}_{\tilde{\alpha}} \gets \exp(\tilde{\alpha})\,\tfrac{1}{d}\sum_{i=1}^d \big(\exp(\tilde{\delta}_i) - \delta^{\text{target}}\big)$; \quad $\tilde{\alpha} \gets \tilde{\alpha} - \eta_{\tilde{\alpha}}\nabla_{\tilde{\alpha}}\mathcal{L}_{\tilde{\alpha}}$
    \State \textit{// Critic update} (Eq.~\eqref{eq:critic loss})
    \For{$k=1,\dots,K_c$}
        \For{each mini-batch of size $B$ sampled from $\{(\mathbf{s}_t^{(n,j)},\mathcal{J}_t^{(n,j)})\}$}
            \State $\phi \gets \phi - \eta_\phi\nabla_\phi\,\tfrac{1}{B}\sum\big(V_\phi(\mathbf{s}) - \mathrm{sg}(\mathcal{J})\big)^2$
        \EndFor
    \EndFor
    \State Polyak update: $\phi' \gets \rho\,\phi' + (1-\rho)\,\phi$
\EndFor
\end{algorithmic}
\end{algorithm}

We also experiment with causality and eligibility traces, where the return in~\eqref{eq: short_horizon_return} is computed as
\begin{equation}
    \mathcal{J} = (1-\lambda)\Big(\sum_{k=1}^{H-t-1}\lambda^k \mathcal{J}_t^k\Big) + \lambda^{H-t-1}\mathcal{J}_t^{H-t},
\end{equation}
analogous to value function learning, i.e., drop past rewards and averaging future consequences.
In practice, however, we find these techniques have negligible effect on training performance.

\subsection{Action bounds}

The action output of the neural network is passed through a $\tanh$ function to satisfy control bounds. However, $\tanh$ saturates for large inputs, reducing the effect of exploration noise; for example, $\tanh(3)\approx0.995$, $\tanh(4)\approx0.999$, and $\tanh(5)\approx0.999$ are nearly identical.
To mitigate this, we clip the pre-activation input before applying $\tanh$. 
In practice, we find that pre-activation clipping plus $\tanh$ yields slightly better behavior than directly clipping without $\tanh$, likely due to the smoother transition provided by $\tanh$.

\subsection{Full algorithm}\label{sec: full algorithm}

Algorithm~\ref{alg: sdpg} summarizes the complete training loop of SDPG with the recommended default
configuration: a state-independent exploration factor $\boldsymbol{\tilde{\delta}}=\log\boldsymbol{\delta}$,
entropy regularization in the actor with automatic temperature tuning, and causal return-to-go with
TD($\lambda$) eligibility traces. 

We use the following notation. The actor $\pi_{\boldsymbol{\theta}}(\cdot\mid\mathbf{o})=\mathcal{N}(\boldsymbol{\mu}_{\boldsymbol{\theta}}(\mathbf{o}),\,\mathrm{diag}(\boldsymbol{\delta})^2)$
outputs a mean from an observation $\mathbf{o}$, while $\boldsymbol{\tilde{\delta}}$ is a state-independent
learnable parameter. The critic $V_\phi$ takes the (privileged) state $\mathbf{s}$, with a target network $V_{\phi'}$
updated by Polyak averaging with coefficient $\rho$. We denote the $N$ nominal environments by indices $n=1,\dots,N$,
and for each nominal $n$ the $M$ auxiliary environments by $(n,j)$ with $j=1,\dots,M$; we use $j=0$ to denote the
nominal trajectory itself.
\section{Additional Results}\label{sec: add results}

\subsection{Eco-centric Training Curve} \label{sec: eco-centric training}
\begin{figure}[t]
    \vspace{-0pt}
    \begin{center}
    \includegraphics[width=\textwidth]{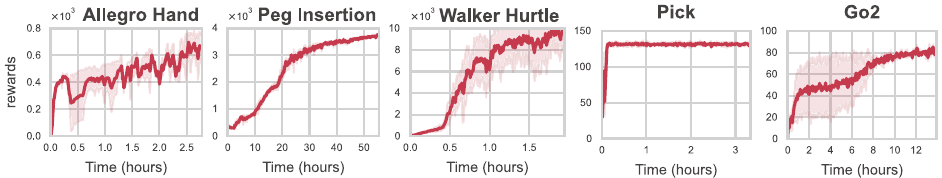}
    \end{center}
    \vspace{-0pt}
    \caption{\small{Example training curve on ego-centric suite.}}
    \vspace{-0pt}
    \label{fig:eco-centric curve}
\end{figure}

Figure~\ref{fig:eco-centric curve} shows additional training curve on ego-centric task suite.

\subsection{Additional Real Experiments} \label{sec: add real exp}

\begin{figure}[h]
    \vspace{-10pt}
    \begin{center}
    \includegraphics[width=\textwidth]{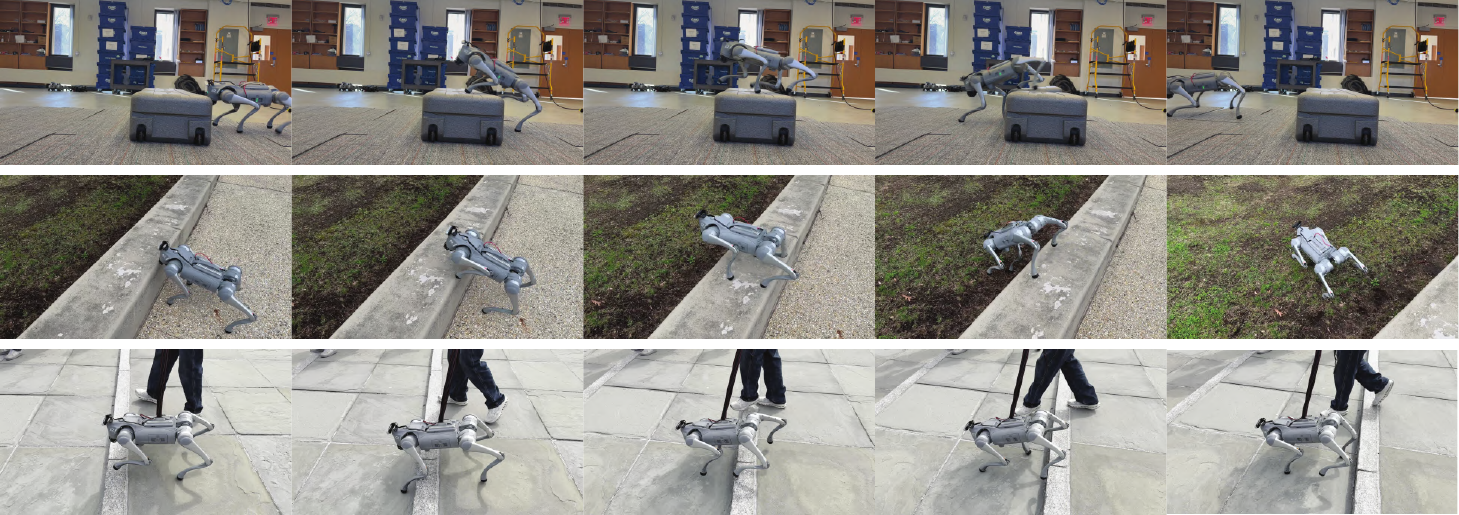}
    \end{center}
    \vspace{-15pt}
    \caption{\small{\textbf{Additional real experiments:} Go2 transpass the box, big stones and down stairs.}}
    \vspace{-0pt}
    \label{fig:add_real_exp}
\end{figure}

Figure~\ref{fig:add_real_exp} show Go2 on other terrian including box, big stones and down stairs.

\section{Additional experiments settings}\label{sec: add exp setting}

\subsection{Environment Settings}\label{sec: env settings}

We evaluate our method on two complementary task suites, both implemented in the GPU-accelerated Genesis simulator:
(i) a \emph{MuJoCo benchmark suite} consisting of four classical continuous control tasks, and
(ii) an \emph{ego-centric tasks suite} that pairs proprioception with on-robot RGB or depth cameras.
The two suites probe complementary aspects of vision-based policy learning:
the MuJoCo benchmark isolates the \emph{visual} component of the problem by giving the actor only third-person images, while the ego-centric suite exercises the more realistic regime in which the policy fuses proprioception with on-board cameras.
Across both suites, we adopt an asymmetric actor--critic setup: the critic always sees the full privileged state of the system, while the actor input is restricted to what we expect to be available at deployment.

\subsubsection{MuJoCo benchmark suite}

This suite consists of four classical control tasks (\textbf{Hopper}, \textbf{Walker}, \textbf{Ant}, \textbf{Humanoid}) following the~\citet{you2025accelerating} setup.
All environments are simulated at $100\,\mathrm{Hz}$ ($\Delta t = 0.01\,\mathrm{s}$) with a maximum episode length of $1000$ steps, and at each reset, the base pose and joint configuration are perturbed with small uniform noise.

\paragraph{Observations}
The privileged state $\textbf{s}_t$ used by the critic follows the SHAC~\cite{xu2022acceleratedpolicylearningparallel} convention and concatenates
(a) base height,
(b) base orientation (quaternion) together with a heading projection toward a forward target,
(c) base linear and angular velocity,
(d) joint positions and (scaled) joint velocities,
and (e) the previous action.
Hopper and Walker are constrained to a 2D sagittal plane and therefore replace the orientation quaternion with a single base pitch angle.
The resulting privileged-state dimensions are $11$, $17$, $37$, and $76$ for Hopper, Walker, Ant, and Humanoid, respectively.

\begin{figure}[h]
    \vspace{-0pt}
    \begin{center}
    \includegraphics[width=\textwidth]{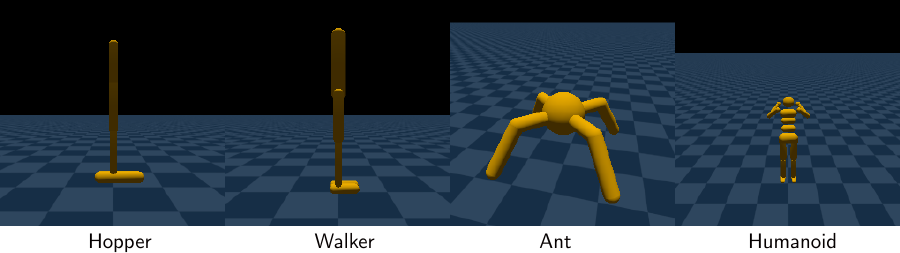}
    \end{center}
    \vspace{-10pt}
    \caption{\small{\textbf{Actor view for Mujoco benchmarks:} The actor receives only a third person view as input.}}
    \vspace{-10pt}
    \label{fig:mujoco_view}
\end{figure}

For vision-based experiments, the actor receives only a stack of the three most recent RGB frames at $84\times 84$ resolution rendered from a third-person camera that tracks the torso of the robot; no proprioception is provided to the actor.
We use a side view for the planar tasks (Hopper, Walker) and a fixed-offset chase view for the spatial tasks (Ant, Humanoid), following~\citet{you2025accelerating}.
Example observations are shown in Figure~\ref{fig:mujoco_view}.

\paragraph{Rewards}
The per-step reward is a weighted sum of task-specific shaping terms that encourage forward locomotion while remaining upright; the terms used by each environment are summarized in Table~\ref{tab: env-rewards}.

\begin{table}[h]
\centering
\small
\caption{Reward terms used in each environment of the MuJoCo benchmark suite. $v_x$ is the base forward velocity, $h$ is the base height, $\theta$ is the base pitch, $R_\text{up}$ and $R_\text{heading}$ are the projections of the base orientation onto the world up and forward axes, and $\mathbf{a}$ is the action.}
\label{tab: env-rewards}
\begin{tabular}{lcccccc}
\toprule
Env & Forward & Height & Angle / Up & Heading & Action penalty & Health \\
\midrule
Hopper   & $v_x$ & $R_h(h)$ & $1 - (\theta/\theta_\text{max})^2$ & --        & $-10^{-1}\|\mathbf{a}\|^2$        & --   \\
Walker   & $v_x$ & --       & --                                 & --        & $-10^{-2}\|\mathbf{a}\|^2$        & $0.1$ \\
Ant      & $v_x$ & $h - h_\text{min}$ & $0.1\,R_\text{up}$        & $R_\text{heading}$ & --                       & --   \\
Humanoid & $v_x$ & $R_h(h)$ & $0.1\,R_\text{up}$                 & $R_\text{heading}$ & $-2\!\times\!10^{-3}\|\mathbf{a}\|^2$ & -- \\
\bottomrule
\end{tabular}
\end{table}

The piecewise height term $R_h(h)$ used by Hopper and Humanoid penalizes states below the nominal standing height much more strongly than it rewards states above it:
\begin{equation}
R_h(h) = \begin{cases}
-200\,\Delta_h^2, & \Delta_h \le 0, \\
\Delta_h,         & \Delta_h > 0,
\end{cases}
\qquad
\Delta_h = \mathrm{clip}(h - h^\star,\, -1,\, h^\star_\text{tol}),
\end{equation}
where $h^\star$ is the nominal base height of each robot and $h^\star_\text{tol}$ is a small tolerance.

\paragraph{Terminations}
Each environment ends the episode when the robot has clearly fallen, leaves the camera frame, or when the simulator state becomes non-physical (e.g.\ due to physics blow-up); the conditions are summarized in Table~\ref{tab: env-terminations}.

\begin{table}[h]
\centering
\small
\caption{Early termination conditions for each environment of the MuJoCo benchmark suite. $h$ denotes the base height (or root $z$ position for the planar tasks) and $\theta$ denotes the base pitch.}
\label{tab: env-terminations}
\begin{tabular}{ll}
\toprule
Environment & Termination condition \\
\midrule
Hopper   & $h < -0.45$, $h > 15$, or $|\dot q| > 100$ on any joint \\
Walker   & $h \notin [-0.5,\, 0.7]$ or $|\theta| > 1.0$ \\
Ant      & $h < 0.27$ or $h > 1.3$ \\
Humanoid & $h < 0.74$ or $h > 2.0$ \\
\bottomrule
\end{tabular}
\end{table}

\subsubsection{Ego-centric tasks suite}\label{sec: ego centric tasks}

\begin{figure}[h]
    \vspace{-0pt}
    \begin{center}
    \includegraphics[width=\textwidth]{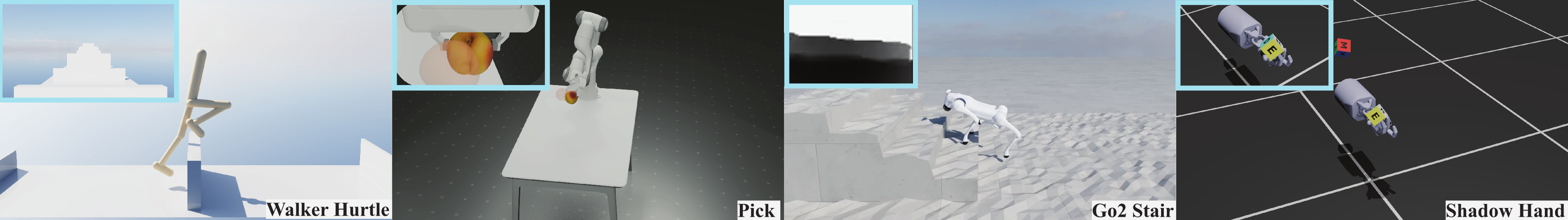}
    \end{center}
    \vspace{-10pt}
    \caption{\small{Additional views for ego-centric tasks.}}
    \vspace{-15pt}
    \label{fig:add_eco_view}
\end{figure}

The ego-centric suite covers a broader set of robots and behaviors that we believe are more representative of how vision-based policies are deployed in the real world: locomotion over obstacles, in-hand manipulation, and bimanual assembly.
The full list of environments is shown in Table~\ref{tab: ego-centric envs}.
For all of these tasks, the actor receives a \emph{combination} of proprioception and on-robot camera observations, and the critic additionally receives the privileged state of any task-relevant external objects (e.g.\ object pose, target pose, contact forces, terrain heightfield).
Example observations are shown in Figure~\ref{fig:add_eco_view}.

\begin{table}[h]
\centering
\small
\caption{Ego-centric tasks suite. ``Modality'' indicates which camera modalities are supported by the environment; ``\#~Cameras'' is the number of cameras whose images are concatenated into the actor input.}
\label{tab: ego-centric envs}
\begin{tabular}{llcc}
\toprule
Environment & Robot / Task & Modality & \# Cameras \\
\midrule
WalkerHurtle    & Planar walker, jump over hurdles            & RGB / Depth & 1 \\
G1Hurtle        & Unitree G1 humanoid, jump over hurdles      & RGB / Depth & 1 \\
Go2Terrain      & Unitree Go2 quadruped on procedural terrain & RGB / Depth       & 1 \\
FrankaPickCube  & Franka Emika Panda, pick an object           & RGB         & 1 \\
AlohaInsertion  & Bimanual ALOHA single-peg insertion         & RGB         & 3 \\
AllegroHand     & In-hand cube reorientation                  & RGB         & 1 \\
ShadowHand      & In-hand cube reorientation                  & RGB         & 1 \\
\bottomrule
\end{tabular}
\end{table}

\paragraph{Proprioception}
Each environment exposes a robot-specific proprioceptive vector that is consistent with what could realistically be measured on the corresponding hardware: projected gravity, base linear and angular velocity, joint positions and velocities, and the user commands.
For the manipulation tasks, the proprioception further includes the gripper / fingertip pose and any task target that is given to the operator (e.g.\ the in-hand goal orientation), but never the pose or velocity of the manipulated object, which the actor must instead infer from the cameras.

\paragraph{Ego-centric cameras}
Cameras are rigidly attached to a link of the robot and follow that link as the robot moves.
The locomotion tasks (WalkerHurtle, G1Hurtle, Go2Terrain) use a single forward-looking camera attached to the base, the manipulation tasks (Franka, ALOHA) use one or more wrist cameras, and the in-hand tasks (Allegro, Shadow) use a fixed external camera that points at the palm.
For locomotion environments, we additionally support a depth-image modality.

\paragraph{Rewards.}
\begin{itemize}[leftmargin=1em]
    \item \textbf{WalkerHurtle} uses the same reward structure as the planar Walker (Table~\ref{tab: env-rewards}) -- a forward-velocity term, an action-magnitude penalty, and a constant health bonus -- but with all three weights scaled up by a factor of $10$ ($10\,v_x$, $-10^{-1}\|\mathbf{a}\|^2$, $+1.0$) so that the per-step return is comparable to the magnitude of an individual hurdle clearance.
    \item \textbf{G1Hurtle} reuses the Humanoid reward of Section~\ref{sec: env settings} (height + forward + up + heading + action penalty), with the height-reward scale $k_h{=}10$, the up-reward scale $0.1$, and the action penalty $-2\!\times\!10^{-3}\|\mathbf{a}\|^2$.
    \item \textbf{Go2Terrain} adopts the velocity-tracking reward used in legged-gym-style locomotion training~\citep{pmlr-v164-rudin22a}: exponential rewards on linear- and angular-velocity command tracking; smoothness penalties on vertical base velocity, off-axis angular velocity, joint acceleration, action rate, and action smoothness; gait-shaping rewards on feet air time, feet clearance, and feet contact while standing; a hip-position regularizer; and safety penalties on joint-limit violation and undesired body contacts.
    \item \textbf{FrankaPickObject} adapts the staged reward of PandaPickCube from IsaacLab~\citep{nvidia2025isaaclabgpuacceleratedsimulation}: a gripper-to-cube reaching term, a cube-to-target tracking term gated on the gripper having reached the cube, a grasp-detection term, a lift bonus, two pose-tracking terms that fire only after the cube is lifted, an arm-posture regularizer, a no-floor-collision term, and a small quadratic action penalty.
    \item \textbf{AlohaInsertion} follows the tolerance-shaped reward of AlohaSinglePegInsertion from MuJoCo Playground~\citep{zakka2025mujocoplayground}: per-arm reaching rewards toward the peg and the socket, per-arm posture regularizers, a no-table-collision term, soft alignment terms gated on the peg and socket being lifted with their $z$-axis pointing up, separate goal-position terms for the peg tip and the socket entrance, and a final peg-insertion reward that is gated on the peg axis being aligned with the socket interior.
    \item \textbf{AllegroHand} (and \textbf{ShadowHand}) adapt the in-hand cube reorientation reward of IsaacGymEnvs~\citep{makoviychuk2021isaacgymhighperformance}: a distance penalty between the cube and the desired in-hand position, a quadratic quaternion-distance rotation reward toward the target orientation, a small action penalty, and a constant healthy bonus that is paid as long as the cube remains in the hand.
\end{itemize}

\paragraph{Terminations.}
Every ego-centric environment ends the episode early as soon as the task has clearly failed or the simulator state becomes unreliable, but the precise condition is task-specific:
\begin{itemize}[leftmargin=1em]
    \item \textbf{WalkerHurtle} uses the same termination as Walker (Table~\ref{tab: env-terminations}).
    \item \textbf{G1Hurtle} terminates when the base height leaves $[0.6,\,1.0]\,\mathrm{m}$ or when the robot drifts past the lateral edge of the terrain.
    \item \textbf{Go2Terrain} terminates when the contact force on the base link exceeds $1\,\mathrm{N}$, or when the absolute roll or pitch of the base exceeds $0.4\,\mathrm{rad}$.
    \item \textbf{FrankaPickCube} terminates when the cube leaves the $\pm 1\,\mathrm{m}$ workspace cube, falls through the table, or when the end-effector dips below the floor.
    \item \textbf{AlohaInsertion} terminates when either the peg or the socket leaves the $\pm 1\,\mathrm{m}$ workspace cube.
    \item \textbf{AllegroHand} and \textbf{ShadowHand} terminate when the cube falls more than $0.2\,\mathrm{m}$ away from the in-hand target position.
\end{itemize}

\subsection{Training Details}\label{sec: training details}

All experiments use the Adam optimizer, a TD($\lambda$) bootstrapped return target with $\gamma{=}0.99$ and $\lambda{=}0.95$, and the same parallel-rollout structure with $N{=}64$ nominal trajectories and $M{=}63$ auxiliary environments per nominal ($N(M{+}1){=}4096$ environments simulated in parallel).

\begin{table}[h]
\centering
\small
\caption{SDPG hyperparameters that are shared across all environments.}
\label{tab: shared hypers}
\begin{tabular}{ll}
\toprule
Hyperparameter & Value \\
\midrule
Nominal trajectories per epoch ($N$)        & $64$ \\
Auxiliary environments per nominal ($M$)    & $63$ \\
Total parallel environments                 & $4096$ \\
Discount factor ($\gamma$)                  & $0.99$ \\
TD($\lambda$) coefficient                   & $0.95$ \\
Critic mini-batch size                      & $4096$ \\
Critic updates per epoch                    & $2$ \\
Actor updates per epoch                     & $1$ \\
Mean action clip                            & $[-2, 2]$ \\
$\Delta J$ normalization                    & enabled \\
Optimizer                                   & Adam \\
Max gradient norm                           & $1.0$ \\
Actor LR schedule                           & cosine, $100$-epoch warmup, $\eta_\text{min}{=}10^{-5}$ \\
Critic LR schedule                          & linear, $1.0\!\to\!0.1$ over training \\
Entropy target std (when used)              & $0.15$ \\
\bottomrule
\end{tabular}
\end{table}

\begin{table}[h]
\centering
\scriptsize
\caption{Per-environment hyperparameters of SDPG in the vision setting. The actor / critic head widths are the widths of the policy / value MLP that sits on top of the per-source encoders described in the text. Go2Terrain is the only environment that uses a non-unit reward scale ($\times 10$) and disables both the causal and eligibility-trace return estimators. AlohaInsertion uses $N{=}48$ nominal trajectories instead of the default $64$ to fit the four image encoders in GPU memory.}
\label{tab: per-env hypers}
\begin{tabular}{lccccccc}
\toprule
Environment & Actor head & Critic head & $H$ & $\eta_\pi$ & $\eta_V$ & $\alpha$ & Ent. \\
\midrule
\multicolumn{8}{l}{\emph{MuJoCo benchmark suite}} \\
Hopper        & $[128,64,32]$        & $[64,64]$           & $32$ & $2\!\times\!10^{-3}$ & $2\!\times\!10^{-4}$ & $0.01$ & yes \\
Walker        & $[128,64,32]$        & $[64,64]$           & $32$ & $2\!\times\!10^{-3}$ & $2\!\times\!10^{-4}$ & $0.01$ & yes \\
Ant           & $[256,64,32]$        & $[64,64]$           & $32$ & $2\!\times\!10^{-3}$ & $2\!\times\!10^{-3}$ & $0.20$ & no  \\
Humanoid      & $[256,128]$          & $[128,128]$         & $16$ & $2\!\times\!10^{-3}$ & $5\!\times\!10^{-4}$ & $0.20$ & no  \\
\midrule
\multicolumn{8}{l}{\emph{Ego-centric tasks suite}} \\
WalkerHurtle    & $[256,128]$          & $[128]$              & $32$ & $1\!\times\!10^{-3}$ & $2\!\times\!10^{-4}$ & $0.01$ & no  \\
G1Hurtle        & $[256,128]$          & $[128]$              & $16$ & $1\!\times\!10^{-3}$ & $5\!\times\!10^{-4}$ & $0.20$ & yes \\
Go2Terrain      & $[512,256,128]$      & $[512,256,128]$      & $24$ & $5\!\times\!10^{-4}$ & $5\!\times\!10^{-4}$ & $0.00$ & no  \\
FrankaPickCube  & $[1024,512,256,128]$ & $[1024,512,256,128]$ & $64$ & $2\!\times\!10^{-3}$ & $5\!\times\!10^{-4}$ & $0.20$ & no  \\
AlohaInsertion  & $[512,256,128]$      & $[512,256,128]$      & $64$ & $2\!\times\!10^{-3}$ & $2\!\times\!10^{-3}$ & $0.20$ & no  \\
AllegroHand     & $[256,128]$          & $[1024,512,256,128]$ & $32$ & $2\!\times\!10^{-3}$ & $5\!\times\!10^{-4}$ & $0.20$ & no  \\
ShadowHand      & $[256,128]$          & $[1024,512,256,128]$ & $32$ & $2\!\times\!10^{-3}$ & $5\!\times\!10^{-4}$ & $0.20$ & no  \\
\bottomrule
\end{tabular}
\end{table}

\paragraph{Network architecture.}
Each input source of the actor and critic is first passed through its own per-source encoder, and the resulting features are concatenated before being fed to the policy / value head.
For state sources the encoder is either a pass-through with running mean--variance normalization or a small one-hidden-layer MLP (of width $256$ for proprioception and $[1024, 128]$ for the proprioception-plus-target vector used by FrankaPickCube and AllegroHand);
for the terrain heightfield used by the legged tasks it is a two-layer MLP of widths $[128, 256]$;
for image sources it is the DrQV2 style~\citep{yarats2021masteringvisualcontinuouscontrol} CNN, with output dimension $128$ for the smaller-actor environments and $256$ for the larger-actor environments (the wrist cameras of AlohaInsertion use a smaller output dimension of $32$).
The policy and value heads are both ELU MLPs whose hidden widths we tune per environment (see Table~\ref{tab: per-env hypers}).
The actor outputs the mean and log-std of a Gaussian policy; the mean is clipped to $[-2, 2]$ and the log-std to a per-task range, reported below.

\paragraph{Hyperparameters shared across environments.}
Table~\ref{tab: shared hypers} lists the algorithmic hyperparameters of SDPG that we keep fixed for every environment.
Critic targets are produced by a Polyak-averaged copy of the critic with task-specific coefficient $\alpha$ (Table~\ref{tab: per-env hypers}); the critic is then trained for two passes per epoch with mini-batches of size $4096$, while the actor takes a single first-order step per epoch.
The per-trajectory return difference $\Delta J$ used in the policy update is normalized across the $M$ auxiliary environments of each nominal, gradient norms are clipped at $1.0$, and we always enable the causality and eligibility-trace described in Section~\ref{sec: additional impl details}.
For environments where the entropy-bonus column of Table~\ref{tab: per-env hypers} is enabled, we additionally regularize the actor toward a target standard deviation of $0.15$ with an automatically tuned temperature initialized at $10^{-2}$.

\paragraph{Per-environment hyperparameters.}

The remaining hyperparameters that vary across environments are listed in Table~\ref{tab: per-env hypers}:
the rollout horizon $H$ used for each policy update, the initial actor and critic learning rates $\eta_\pi$ and $\eta_V$, the target-critic Polyak coefficient $\alpha$, the actor / critic head widths, and an ``Ent.'' flag indicating whether the entropy regularizer described above is used.
The log-std of the policy is clipped to $[-5, 2]$ for all environments except Humanoid, where we use the tighter $[-2.2, 0]$ range.

\end{document}